\documentclass[final,5p,times,twocolumn]{elsarticle}
\usepackage{multirow}

\usepackage{amssymb}
\usepackage{latexsym}

\usepackage{url}
\usepackage{xcolor}
\usepackage{hyperref}
\usepackage{graphicx}
\usepackage{amsmath}
\usepackage{upgreek}
\usepackage{array}
\usepackage{graphbox}  
\usepackage{verbatim}  
\usepackage{booktabs}
\usepackage{siunitx}
\usepackage{lineno}
\usepackage{subcaption}
\usepackage{mathtools}
\usepackage{textcomp}
\usepackage{pifont}
\usepackage{svg}

\hypersetup{colorlinks = true,linkcolor = blue,anchorcolor =red,citecolor = blue,filecolor = red,urlcolor = red, pdfauthor=author}

\newcolumntype{C}[1]{>{\centering\let\newline\\\arraybackslash\hspace{0pt}}m{#1}}

\DeclareMathOperator*{\argmax}{max}
\DeclareMathOperator*{\argmin}{min}
\DeclarePairedDelimiter\abs{\lvert}{\rvert}
\newcommand{\mr}[1]{\mathrm{#1}}
\newcommand{\matr}[1]{\mathbf{#1}}
\newcommand{\norm}[1]{\left\|#1\right\|}
\newcommand{\tr}[1]{{#1}^\top}
\newcommand{\yy}{\matr{y}}
\newcommand{\xx}{\matr{x}}
\newcommand{\Loss}{\mathcal{L}}
\newcommand{\Lcyc}{\Loss_{\mathrm{cyc}}}

\newcommand{\Lgan}{\Loss_{\mathrm{LSGAN}}}
\newcommand{\Lprior}{\Loss_{\mathrm{prior}}}
\newcommand{\Id}{\mathrm{Id}}
\newcommand{\EE}{\mathbb{E}}

\newcommand{\logId}{\log_{\Id}}
\newcommand{\expId}{\exp_{\Id}}
\newcommand{\spdMan}{\mathcal{S}_{++}^*}
\newcommand{\spdnMan}[1]{\mathcal{S}_{++}^#1}

\newcommand{\XHR}{X_{\mathrm{HR}}}
\newcommand{\YHR}{Y_{\mathrm{HR}}}
\newcommand{\YLR}{Y_{\mathrm{LR}}}
\newcommand{\genX}{G_X}
\newcommand{\genY}{G_Y}
\newcommand{\disX}{D_X}
\newcommand{\disY}{D_Y}

\newcommand{\phm}{\phantom{0}}

\newcommand{\Prob}{\mathbb{P}}
\newcommand{\PP}{\matr{P}}

\newcommand{\UU}{\matr{U}}
\newcommand{\Eig}{\boldsymbol{\Upsigma}}
\newcommand{\real}{\mathbb{R}}

\newcommand{\nSphere}[1]{\mathbb{S}^{#1}}
\newcommand{\twoSphere}{\mathbb{S}^2}
\newcommand{\cc}{\mathbf{c}}
\newcommand{\vc}{\mathbf{v_c}}
\newcommand{\uu}{\mathbf{u}}
\newcommand{\logU}{\log_{\uu}}
\newcommand{\expU}{\exp_{\uu}}
\newcommand{\Tu}{\mathcal{T}_{\uu}}
\newcommand{\probODF}[1]{p(\,\cdot\,|\,#1)}
\newcommand{\intTwoSphere}{\int_{\mathbf{s} \in \twoSphere}}
\newcommand{\pdf}{\mathcal{P}}
\newcommand{\ltwoMetric}{\mathbb{L}^2}

\newcommand{\Tid}{\mathcal{T}_{\Id}}
\newcommand{\Ss}{\matr{S}}

\newcommand{\KK}{\matr{K}}
\newcommand{\sym}[1]{{#1}_{sym}}
\newcommand{\diag}[1]{{#1}_{diag}}
\newcommand{\dLdM}{\frac{\partial{L^{(k)}}}{\partial{\matr{M}_{k-1}}}}
\newcommand{\dLdU}{\frac{\partial{L^{(k^{'})}}}{\partial{\UU}}}
\newcommand{\dLdE}{\frac{\partial{L^{(k^{'})}}}{\partial{\matr{\Sigma}}}}
\newcommand{\dLdMk}{\frac{\partial{L^{(k+1)}}}{\partial{\matr{M_{k}}}}}

\newcommand{\down}{\mathcal{F}}

\newcommand{\cmark}{\ding{51}}
\newcommand{\xmark}{\ding{55}}

\newcommand{\equaref}[1]{Eq.~\eqref{#1}}
\newcommand{\figref}[1]{Figure~\ref{#1}}
\newcommand{\tabref}[1]{Table~\ref{#1}}
\newcommand{\secref}[1]{Section~\ref{#1}}

\newcommand{\FA}{FA$\!^*$}

\definecolor{newcolor}{rgb}{.8,.349,.1}

\begin{document}

\begin{frontmatter}

\title{Manifold-aware Synthesis of High-resolution Diffusion from Structural Imaging}

\author[1]{Benoit Anctil-Robitaille\corref{cor1}}
\cortext[cor1]{Corresponding author: 
  benoit.anctil-robitaille.1@ens.etsmtl.ca}
\author[2]{Antoine Théberge}
\author[2]{Pierre-Marc Jodoin}
\author[2]{Maxime Descoteaux}
\author[1]{Christian Desrosiers}
\author[1]{Hervé Lombaert}

\address[1]{Department of Computer and Software Engineering, ETS Montreal, Canada}
\address[2]{Department of Computer Science, Sherbrooke University, Canada}

\begin{abstract}
%%%
 The physical and clinical constraints surrounding diffusion-weighted imaging (DWI) often limit the spatial resolution of the produced images to voxels up to 8 times larger than those of T1w images. Thus, the detailed information contained in T1w images could help in the synthesis of diffusion images in higher resolution. However, the non-Euclidean nature of diffusion imaging hinders current deep generative models from synthesizing physically plausible images. In this work, we propose the first Riemannian network architecture for the direct generation of diffusion tensors (DT) and diffusion orientation distribution functions (dODFs) from high-resolution T1w images. Our integration of the Log-Euclidean Metric into a learning objective guarantees, unlike standard Euclidean networks, the mathematically-valid synthesis of diffusion. Furthermore, our approach improves the fractional anisotropy mean squared error (FA MSE) between the synthesized diffusion and the ground-truth by more than 23\% and the cosine similarity between principal directions by almost 5\% when compared to our baselines. We validate our generated diffusion by comparing the resulting tractograms to our expected real data. We observe similar fiber bundles with streamlines having less than 3\% difference in length, less than 1\% difference in volume, and a visually close shape. While our method is able to generate high-resolution diffusion images from structural inputs in less than 15 seconds, we acknowledge and discuss the limits of diffusion inference solely relying on T1w images. Our results nonetheless suggest a relationship between the high-level geometry of the brain and the overall white matter architecture.
%%%%
\end{abstract}


\end{frontmatter}

\section{Introduction}

Diffusion MRI is of crucial importance in multiple challenging tasks, including the diagnosis of complex cognitive disorders \citep{Kelly2018,Kantarci2017,Neuner2010}, the study of neurodegenerative diseases \citep{huang2007diffusion, gattellaro2009white} and neurosurgical planning \citep{Costabile2019}. Nonetheless, diffusion-weighted imaging (DWI) suffers from a low signal-to-noise ratio (SNR) and a poor spatial resolution arising from physical and clinical limitations such as the use of echo-planar imaging (EPI) and limited patient scanning time. Indeed, the induced trade-off between image resolution, SNR and imaging time \citep{Poot2013} in the acquisition of DWI often results in images with voxel size up to 8 times larger than other common modalities such as structural T1w images, e.g. 1 mm iso. for DWI vs 2 mm iso. for T1w \citep{Shi2016}. Hence, it has been shown that the increased presence of partial volume effect (PVE) in DW images, due to their low spatial resolution, impairs their subsequent analysis \citep{Oouchi2007, alexander2001analysis}.

%%Existing methods for fast high-resolution diffusion imaging falls into two main categories: 1) design and tuning of specialized hardware and sequences, and 2) images reconstruction and post-processing. While the development of better hardware and specialized acquisition sequences is fundamental and has recently made great progress \citep{Holdsworth2019}, it cannot help improve already acquired images and is of limited clinical accessibility. The second family of methods aims at reconstructing high-resolution (HR) images from pre-existing low-resolution (LR) ones, which favors an easier adoption and a quicker impact. 

These limitations have triggered the development of post-processing methods that aim to improve the spatial resolution of low-resolution diffusion volumes. Tackling the estimation of high-resolution (HR) diffusion from low-resolution (LR) images was first explored with interpolation-based methods \citep{Arsigny2006,Yap2014,Dyrby2014}. They resample existing images to a higher-resolution grid and is nowadays a default step in diffusion MRI processing tools such as in TractoFlow \citep{Theaud2020} and MRtrix3 \citep{Tournier2019}. Although fine anatomical details can be enhanced with such approach, interpolations will always be limited by the inherent coarseness of the original diffusion data as they exclusively rely on intra-image information.

Machine learning offers an effective way to leverage the rich information contained in HR images for the synthesis of diffusion imaging in the same resolution, thus going beyond interpolation. In \citep{Alexander2017}, a fully supervised image quality transfer (IQT) framework using random forests is proposed to learn a non-linear mapping between paired low-quality and high-quality diffusion data. Similarly, in \citep{Elsaid2019}, a supervised 2D SRCNN is used for the same objective. The authors demonstrate that learning a mapping from an LR input to its HR version not only helps recovering anatomical details better than interpolation, but can also help in downstream tasks such as tractography. However, such approaches rely on limited high-resolution diffusion data which are costly and challenging to acquire. Moreover, the methods in \citep{Alexander2017,Elsaid2019} have only been tested on small datasets comprising a maximum of 23 subjects and 3 subjects respectively.

In parallel, deep neural networks offer unsupervised learning techniques that only require few paired training samples to train specific synthesis tasks. More particularly, Generative Adversarial Networks (GANs) \citep{Goodfellow2014} have been successfully used for the synthesis of missing modalities \citep{Dar2019}, image-to-image translation \citep{Zhu2017, Lei2019} and image super-resolution \citep{sanchez2018brain}, just to name a few. Therefore, deep generative models could be a key solution for the synthesis of high-resolution diffusion from unpaired images expressing a higher level of structural details such as T1w images.

While the synthesis of raw DWI signals with a proper angular resolution is an extremely resource-intensive task, current deep learning architectures struggle to generate plausible diffusion reconstruction schemes, notably Diffusion Tensor (DT) or Orientation Distribution Functions (ODFs) because of their non-Euclidean nature \citep{Huang2019}. Indeed, each voxel of a DT image lies on a Riemannian manifold of symmetric positive definite (SPD) 3$\times$3 matrices \citep{Arsigny2006}, and ODFs can be represented as points on an n-Sphere manifold $\nSphere{n}$ \citep{Cheng2009}. In the context of image synthesis, the inability of networks to capture the underlying non-linear Riemannian manifold geometry of the data results in the generation of implausible images that miss the important mathematical properties of diffusion imaging \citep{Huang2019}. Consequently, the limitations of current deep neural networks (DNN) have impeded the development of generative models in diffusion imaging, which have been mostly restricted to the synthesis of DT scalar maps such as Fractional Anisotropy (FA) and Mean Diffusivity (MD).

\begin{figure}[t!]
\centering
\includegraphics[width=.95\linewidth]{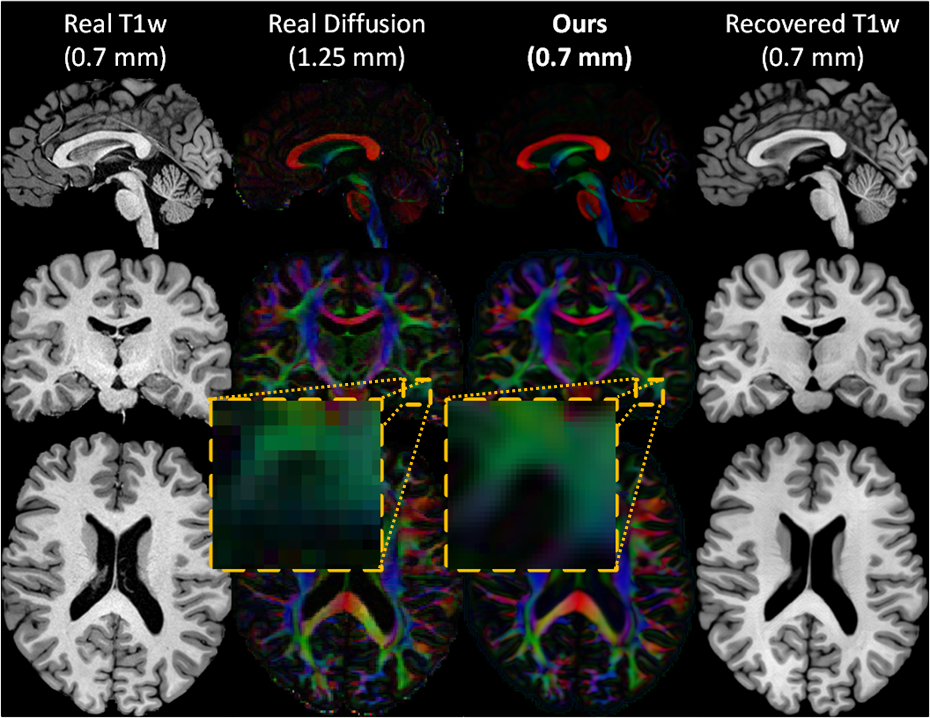}
\caption{An example of the input and output of our network. The HR T1w input image (\textbf{first column}), the real LR diffusion (\textbf{second column}), the generated HR diffusion (\textbf{third column}) and the recovered HR T1w image (\textbf{last column}) of a test subject.} \label{fig:odf_forward_cycle_color_FA}
\end{figure}

\subsection{Related Work}
\paragraph{Structural-to-Diffusion Synthesis}
Amidst the literature, \citep{Gu2019} studies the generation of diffusion-derived scalar maps from downsampled structural images. To do so, the authors use a CycleGAN to learn the intermodal relationships between T1w images and FA/MD maps and successfully translate one to another. They demonstrate that structural images share sufficient information with the diffusion anisotropy of tissues to synthesize plausible 2D FA and MD slices. Similarly, in \citep{Lan2020}, a Self-attention Conditional GAN (SC-GAN) is used to generate FA and MD maps from different input modalities including structural T1w images. Their results indicate that both the 3D contextual information and the adversarial objective are important building blocks for the synthesis of diffusion data. In \citep{Zhong2020}, dual GANs with a Markovian discriminator \citep{Li2016} are employed for the harmonization of inter-site DT-derived metrics. Finally, in \citep{Son2019} functional MRI in combination with structural T1w inputs are fed to a CNN network to generate DT. 
This body of work demonstrates the potential of generative models for the structural-to-diffusion synthesis of imaging data. Nevertheless, they remain limited \citep{Gu2019, Lan2020, Son2019} in not exploiting the high-resolution information contained in the structural images to their full extent by either considering downsampled version of the T1w inputs or 2D slices with limited context. In addition, even though diffusion scalar maps are clinically useful, they mostly ignore fiber orientations and are of limited interest for tasks such as tractography or connectome visualization. With regards to the generated tensors in \citep{Son2019}, the authors provide no guarantee on their mathematical validity, such as symmetric positive definiteness, nor on their usability in a downstream task such as tractography.

%\paragraph{Guided Super-Resolution}
%Guided super-resolution aims to improve the quality of an image acquired with a modality A using a corresponding high-quality guide image in a second modality B that is cheaper to acquire at the desired level of details. This concept, that has yet to be developed in the medical imaging field, has shown promising results in the computer vision community and is a key concept of our proposed framework. For instance, guided super-resolution has been widely used to learn the upsampling of costly depth maps using corresponding cheap HR RGB images \citep{Riegler2016, Ni2017, Song2017, Li2019, DeLutio2019,Song2020}. These works demonstrate the ability of deep learning models to transfer high-frequency information from an HR guide image to an LR counterpart sharing similar structures through modalities. Using elements of image super-resolution techniques such as pre-upsampling \citep{Riegler2016, Song2017, Ni2017, Li2019}, residual learning and attention mechanisms \citep{Song2020} to merge paired source and HR guide images, the authors are able to considerably increase the level of detail of the LR source images. Unfortunately, these works strictly rely on supervised training which limits their application to real world scenarios where the HR ground-truths are hardly accessible or nonexistent. Analogously, our work proposes to use T1w images as guides to synthesize DT and ODFs in high-resolution using unpaired samples and limited priors by extending the CycleGAN architecture \citep{Zhu2017} as detailed in \figref{fig:architecture}.

\paragraph{Manifold-Valued Data Learning}
Deep learning models are well suited to model data lying in an Euclidean vector space. However, the Euclidean operations from which they are built upon, e.g., convolutions or pooling, are not well defined on curved manifolds. Moreover, the application of Euclidean geometry to manifold-valued data, such as DT and ODF, has well-documented side effects \citep{Arsigny2006}. Consequently, studies that use neural networks for the accurate processing of data on Riemannian manifolds have started to emerge \citep{Chakraborty2019, brooks2019riemannian}. However, these works require substantial modifications of known deep learning models and call for further investigation in a broader set of scenarios. 

Another avenue for the processing of manifold-valued data resides in the design of computationally efficient Riemannian metrics. To that purpose, \citep{Arsigny2006} proposed a Log-Euclidean metric to process data lying on the $\spdMan$ manifold with applications to diffusion tensors. With the help of the $\log$ and $\exp$ maps defined in \citep{Arsigny2006}, one can process tensors using Euclidean operations and guarantee that the processed tensors keep their SPD properties. Likewise, a Log-Euclidean framework has also been proposed in \citep{Cheng2009} for the computation of orientation distribution function and applied to diffusion ODF. These two frameworks, combined with the matrix backpropagation of spectral layers presented in \citep{Ionescu}, constitutes the fundamentals of the following manifold-valued data learning approaches. For instance, in \citep{Huang2016}, the authors have integrated the Log-Euclidean metric into their deep learning model called SPDNet to learn compact and discriminative SPD matrices. Although SPDNet offers a way to learn data on $\spdMan$, it has not been designed for spatially organized and volumetric SPD matrices learning as in DT.

More recently, \citep{Huang2019} proposed a Wasserstein GAN (WGAN) \citep{Arjovsky2017} leveraging the Log-Euclidean metric to synthesize plausible DT, among other manifold-valued data type. To ensure the validity of the generated data, the authors project the output of their generator network to a vector space using the $\log$ map from \citep{Arsigny2006} prior to the discriminator assessment. The $\exp$ operation is then used on the synthesized output to recover valid DT. Despite its ability to generate mathematically valid diffusion, the model in \citep{Huang2019} outputs images that are not conditioned by any real subject specific information (e.g., a T1w image) and, thus, are less clinically valuable. In addition, this prior work only focuses on the generation of DT in 2D, which once again limits the value of the generated data.

\subsection{Contributions}

This work proposes a novel deep learning architecture that leverages the detailed information of high-resolution structural images to guide the synthesis of DT and ODF in the same high-resolution space. Based on the CycleGAN architecture \citep{Zhu2017}, our solution exploits the inherent cross-modality representations of structural and diffusion images to learn functions that map one to the other in a weakly supervised manner. To do so, we address the current limitations of deep learning models built upon Euclidean operations by integrating a Riemannian framework, namely the Log-Euclidean framework, for statistical computations on DT and ODF directly into the model. Such Riemannian framework within the learning procedure of the network enforces a valid synthesis of diffusion data that lies on a desired Riemannian manifold. By constraining the generated diffusion to lie on the Riemannian manifold of $3\times3$ symmetric positive definite matrices ($\spdnMan{3}$) for DT and on the n-Sphere $\nSphere{\mathbf{n}}$ manifold for ODF, we guarantee the mathematical coherence of the solution. As opposed to existing deep generative models that focused on the generation of diffusion derived scalar maps \citep{Li2016, Gu2019, Lan2020}, our model outputs complete diffusion schemes, here the $3 \times 3$ DT and the spherical harmonic coefficients of ODF. This important difference allows one to perform tractography, tractogram visualization and fiber bundle segmentation in addition to scalar maps computation directly from our network output while only requiring a T1w image as input. 

Specifically, our contributions are as follows:
\begin{itemize}
  \item The first Riemannian network for the cycle-consistent mapping between real-valued images and data lying on the $\spdnMan{3}$ and the $\nSphere{\mathbf{n}}$ manifolds;
  
  \item The first deep learning model for the guided super-resolution of DT and ODF from unpaired high-resolution structural images and limited priors;
  
  \item A comprehensive analysis of synthesized diffusion imaging including the evaluation of full-valued diffusion data, scalar maps and tractography.
\end{itemize}

%The following sections of our manuscript is organised as follows. \textcolor{red}{TO BE COMPLETED !!}

\section{Method}\label{section:Method}

\begin{figure*}[t!]
\centering
\includegraphics[width=.70\linewidth]{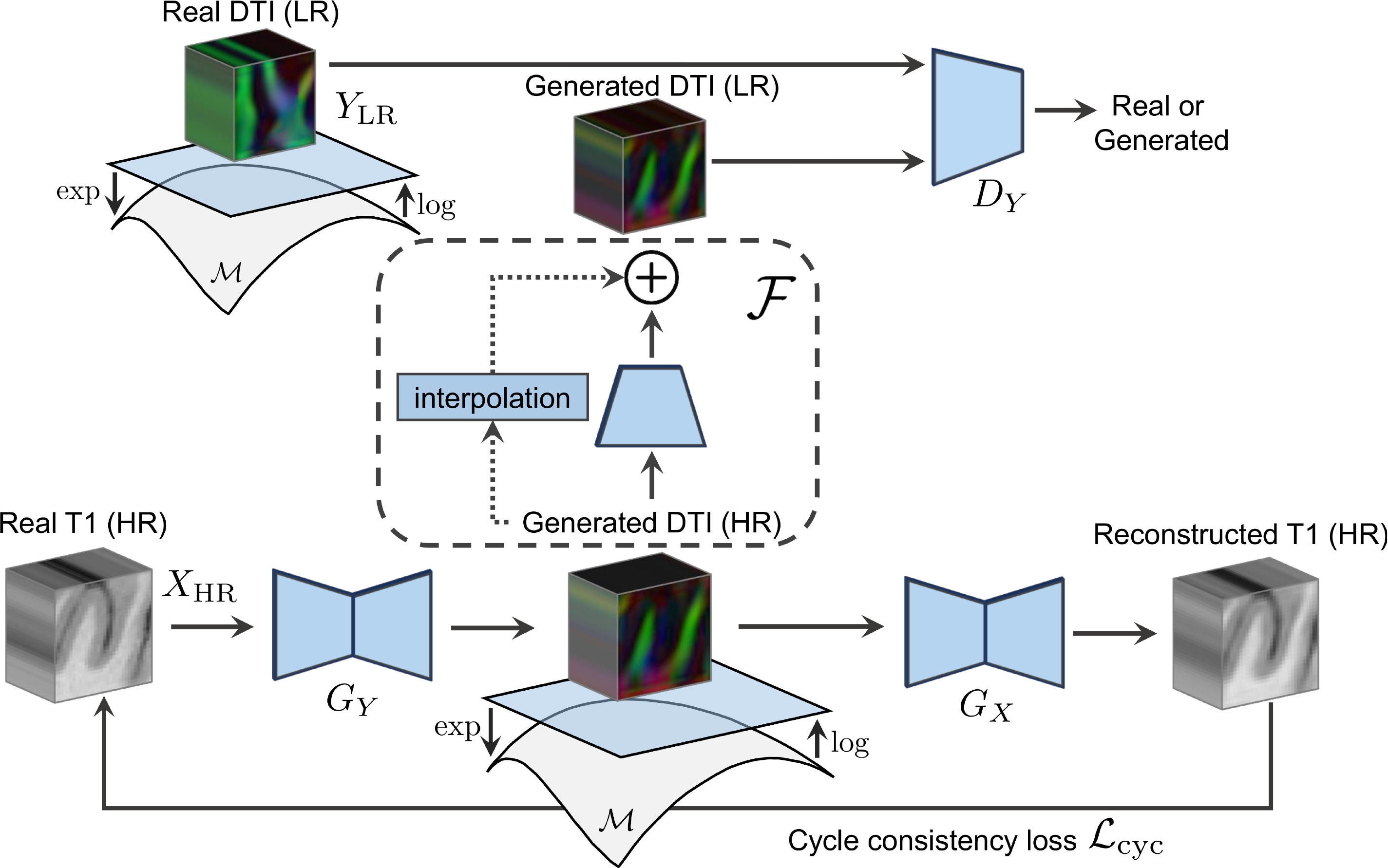}
\caption{The forward cycle of our manifold-aware CycleGAN. $G_Y$ generates high resolution DT on the $\spdnMan{3}$ manifold using the $\logId$ and $\expId$ mapping. $D_Y$ assesses the generated images quality and provides feedback to $G_Y$. $G_X$ tries to reconstruct the original T1w images from $\logId(\expId(G_Y(\xx)))$.} \label{fig:architecture}
\end{figure*}

Let $X$ be the real-valued domain of structural images (e.g., T1w) and $Y$ be the manifold-valued domain of diffusion images (e.g., DT or ODF). We aim at learning mapping functions $\genY: \XHR \mapsto \YHR$ and $\genX: \YHR \mapsto \XHR$ that translate high-resolution T1w images to high-resolution diffusion images and the other way around. Nevertheless, we mainly have access to unpaired HR T1w images and LR diffusion data that differ in terms of subject, resolution and nature. To address this problem, we propose a Manifold-Aware CycleGAN (MA-CycleGAN) architecture (see \figref{fig:architecture}) that inherently handles both the domain translation and the super-resolution of diffusion, while accounting for the Riemannian geometry of the data. 

We train our network with unpaired training samples $\{\xx_{i}\}^{N}_{i=1}$ where $\xx_{i} \in \XHR $ is a 3D structural image, and $\{\yy_{j}\}^{M}_{j=1}$ where $\yy_{j} \in \YLR$ is a diffusion image in a lower spatial resolution. The mathematical validity of the generated DT and ODF is ensured by projecting every synthesized voxel on the tangent plane at their isotropic counterpart (i.e., the $3\times3$ identity matrix for DT and the uniform distribution for ODF). To do so, the exponential and logarithm maps borrowed from their underlying Riemannian manifolds geometry are used. Two discriminators $\disX$ and $\disY$ evaluate the synthesized HR T1w images $\genX(\yy)$ and downsampled HR diffusion images, using a learned residual function $\down$ as follows $\down(\genY(\xx))$, with regards to their real data distribution $\genX(\yy)\sim{\Prob_{\XHR}}$ and $\down(\genY(\xx)) \sim{\Prob_{\log(\YLR)}}$. By combining pixel-wise reconstruction losses and higher-level adversarial feedback in a single objective, our MA-CycleGAN is able to exploit the local tissue information and global geometry of high-resolution structural images to produce realistic and usable sharp diffusion data. 

In the following sections, we detail the Riemannian frameworks embedded in our architecture for DT and ODF learning. Moreover, we frame our cycle-consistent and adversarial objectives incorporating both the $\exp$ and $\log$ maps of the aforementioned Riemannian frameworks and an up-and-down sampling strategy. Then, we present our anisotropy-based attention mechanism that helps the network to focus on meaningful fiber tracts information.

\subsection{Riemannian Framework for Diffusion Tensors Learning}\label{sec:dti_riemannian_framework}

Diffusion tensors are $3 \times 3$ symmetric positive matrices $\matr{M}$ that can be decomposed in three real and positive eigenvalues $\Eig$ and three corresponding eigenvectors $\UU$ using eigendecomposition such that $\matr{M} = \UU \Eig \tr{\UU}$. The eigenvector $\matr{u_1}$ associated with the largest eigenvalue of $\matr{M}$ represents the principal direction of diffusion and aligns with the underlying fibers population. DT-derived metrics, describing the shape of the tensor, are computed from the positive eigenvalues $\lambda_1 > \lambda_2 > \lambda_3 \in \Eig$. One of the most important DT-derived metric, fractional anisotropy, measures how far the shape of the diffusion tensor is from a sphere (i.e., how anisotropic the diffusion is). This metric is computed as follows:
\begin{equation}\label{eq:fractional_anisotropy}
\begin{split}
   \mathrm{FA}=\sqrt{\frac{1}{2}}\frac{\sqrt{(\lambda_1-\lambda_2)^2+(\lambda_2-\lambda_3)^2+(\lambda_1-\lambda_3)^2}}{\sqrt{(\lambda_{1}^2 + \lambda_{2}^2 + \lambda_{3}^2)}}.
 \end{split}
\end{equation}
Because of their SPD properties, tensors lie on a non-linear manifold denoted as
\begin{equation}
    \spdnMan{3} = \big\{\matr{M} \in \real^{3 \times 3}, \, \matr{M} = \tr{\matr{M}}, \, \tr{\matr{x}}\matr{M}\matr{x} > 0, \, \forall \matr{x} \in \real^3, \, \norm{\matr{x}}_2 > 0\big\} \label{eq:spd_space} \\
\end{equation}
$\spdnMan{3}$ is not a vector space (i.e., the linear combination of two elements in $\spdnMan{3}$ may lie outside this space), thus using standard Euclidean operations to process statistics on diffusion tensors can lead to undesirable effects like the well-documented swelling effect in \citep{Arsigny2006}. Moreover, not considering the $\spdnMan{3}$ manifold while synthesizing DT with deep neural network can lead to the generation of non-SPD tensors as can be seen in \tabref{tab:non_manifold_count}. Such tensors are physically incorrect and must be avoided. To accurately process DT, \citep{Arsigny2006} proposed a Log-Euclidean metric that will be presented in the following sections.   

\begin{figure}[t!]
\centering
\includegraphics[width=.98\linewidth]{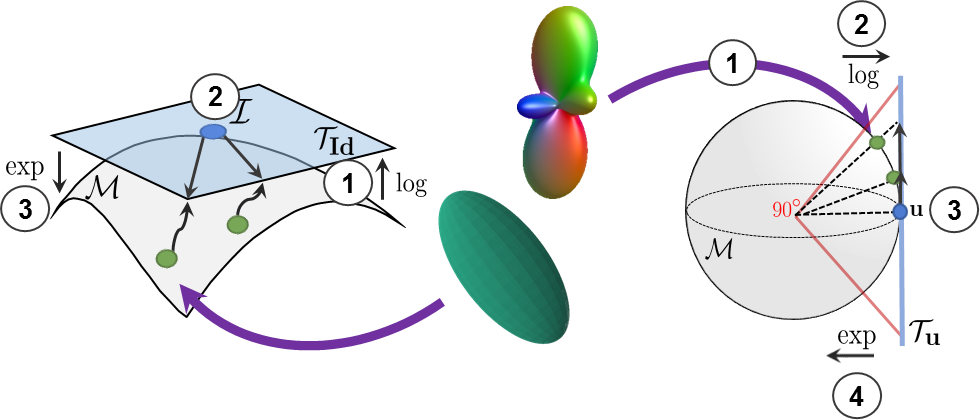}
\caption{We use the Log-Euclidean metric to project the manifold-valued diffusion data to a tangent plane before processing them with Euclidean operations. \textbf{On the left}, the projection of diffusion tensors from the $\spdnMan{3}$ manifold to the tangent plane at the isotropic tensor $\Tid$. \textbf{On the right}, the projection of square-root re-parameterized ODFs from the $\nSphere{n}$ manifold to the tangent plane at the uniform distribution $\Tu$.} \label{fig:log_euclidean_framework}
\end{figure}

\subsubsection{Log-Euclidean Metric}\label{sec:dti_log_euclidean_metric}

Diffusion tensor matrices are well defined in the Log-Euclidean metric, where a matrix logarithm and exponential can be conveniently processed in a metric and can always be mapped back to a valid symmetric diffusion tensor \citep{Arsigny2006}. Let $\matr{M} = \UU \Eig \tr{\UU}$ be the eigendecomposition of a symmetric matrix $\matr{M}$. The computation of the logarithm and the exponential of a tensor noted as $\logId$ and $\expId$ are defined as follows:
\begin{align}
    \forall \PP \in \spdnMan{3}, \ \logId(\PP) & \ = \ \UU\, \log(\Eig)\,\tr{\UU} \, \in \, \Tid\label{eq:log_of_tensor} \\
    \forall \Ss \, \in \, \Tid, \ \expId(\Ss) & \ = \ \UU\,\exp(\Eig)\,\tr{\UU} \, \in \,  \spdnMan{3}\label{eq:exp_of_tensor}
\end{align}
Using these definitions, the geodesic distance between two points on the $\spdnMan{3}$ manifold, $\PP_1$ and $\PP_2$, can then be expressed as
\begin{align}
   dist(\PP_1, \PP_2) \, = \, \big\|\logId(\PP_1)-\logId(\PP_2)\big\|_2\label{eq:spd_log_distance}.
\end{align}

We provide in \ref{appendix_backprop} the definition of the matrix backpropagation for diffusion tensors learning used to train our network and the link to our open-source implementation. For more details, the reader is referred to \citep{Ionescu}. 

\subsection{Riemannian Framework for ODF Learning}

Orientation distribution functions, here represented as $\mathbf{p}(\mathbf{s})$, are probability density functions modelling the diffusion of water molecules at any angle $\mathbf{s}$ on the 2-sphere $\twoSphere$. The space $\mathcal{P}$ of such PDFs forms the set 
\begin{equation}
   \mathcal{P} = \big\{\mathbf{p}: \twoSphere \to \mathbb{R}^+ \, | \, \forall \mathbf{s} \in \twoSphere, \mathbf{p}(\mathbf{s}) \geq 0; \, \intTwoSphere \mathbf{p}(\mathbf{s})d\mathbf{s}=1 \big\} 
\end{equation}
The constrained function space $\pdf$ is not a vector space but a nonlinear differentiable manifold that, just like the aforementioned $\spdnMan{3}$ manifold, needs to be equipped with an efficient Riemannian metric to accurately process statistics on it \citep{Srivastava2007}. Fortunately, PDFs can be re-parameterized in multiple ways leading to known manifolds with closed-form and computationally-efficient Riemannian operations. The square-root re-parameterization of PDFs is a particularly convenient one as it results in a unit Hilbert sphere manifold with an $\ltwoMetric$ metric \citep{Srivastava2007}. 

\subsubsection{Square Root Re-Parameterization of ODF}\label{sec:odfs_reparameterization}

With the help of the square-root re-parameterization of ODF $\psi(\mathbf{s}) = \sqrt{\mathbf{p}(\mathbf{s})}, \forall \mathbf{s} \in \twoSphere$, the space $\boldsymbol{\psi}$ can be viewed as the positive orthant of a unit Hilbert sphere:
\begin{equation}
    \boldsymbol{\psi} = \big\{\psi: \mathbb{S}^2 \to \mathbb{R}^+ \, | \, \forall \mathbf{s} \in \mathbb{S}^2, \psi(\mathbf{s}) \geq0; \, \int_{\mathbf{s} \in \mathbb{S}^2} \psi^2(\mathbf{s})d\mathbf{s}=1 \big\}
\end{equation}
where the geodesic, exponential and logarithm maps are defined in a closed form. In practice, ODFs are represented as histograms of $N$ bins where each bin represents an orientation on a discretized 2-sphere. In the context of learning spatially organized ODFs in 3D, working directly with the re-parameterized ODFs $\psi(\mathbf{s})$ would require considerable resources. Indeed, $N$ is typically in the order of few hundreds. To alleviate this burden, $\psi(\mathbf{s})$ can be represented in a more compact form using a spherical harmonic basis \citep{Descoteaux2007, Cheng2009} as follows:
\begin{equation}\label{eq:basis_representation}
    \psi(\mathbf{s}) = \sum_{i=1}^{K}c_i\mathbf{B}_i(\mathbf{s})
\end{equation}
Here $K \ll N$ is the number of orthonormal basis functions used to represent $\psi(\mathbf{s})$ and $\{\mathbf{B}_i\}_{i \in K}$ is the set of spherical harmonic basis functions as in \citep{Descoteaux2007}. From this parametric representation, an efficient Log-Euclidean framework has been proposed in \citep{Cheng2009} and is presented in \secref{sec:odf_log_euclidean_framework}. Similarly to the FA of the diffusion tensor model, the generalized fractional anisotropy (GFA) \citep{Tuch2004} of the ODF can be computed from the spherical harmonic coefficients as in \equaref{eq:generalized_fractional_anisotropy} below:
\begin{equation}\label{eq:generalized_fractional_anisotropy}
\begin{split}
   \mathrm{GFA}=\sqrt{1-\frac{(c_{0}^0)}{\sum_{k=0}^L\sum_{m=-k}^k(c_{k}^m)^2}}.
 \end{split}
\end{equation}
Here, the GFA measure how far is the ODF from the uniform distribution.

\subsubsection{Log-Euclidean Metric}\label{sec:odf_log_euclidean_framework}

Given the parametric representation of $\psi(\mathbf{s})$ in \equaref{eq:basis_representation}, the square root of any ODF can be expressed by its Riemannian coordinate $\cc=(c_1,c_2,\ldots,c_K)^{\top}$ and gives the probability family $PF_K$:
\begin{equation}\label{eq:odf_probability_family}
\footnotesize
   PF_K=\Big\{p(\mathbf{s}\,|\,\cc)=\Big(\sum_{i=1}^{K}c_i\mathbf{B}_i(\mathbf{s}) \Big)^2: \int_{\mathbf{s}} p(\mathbf{s}\,|\,\cc)d\mathbf{s}=\sum_{i=1}^{K}c_{i}^{2}=1, \sum_{i=1}^{K}c_i\mathbf{B}_i(\mathbf{s}) \ge 0, \forall \mathbf{s} \in \twoSphere\Big\}.
\end{equation}
Following \equaref{eq:odf_probability_family}, the parameter space $PS_K$ can be defined as
\begin{equation}\label{eq:odf_parameter_space}
   PS_{\!K} = \big\{ \cc \ | \, \norm{\cc} = \sum_{i=1}^K c_i^2 =1, \ \sum_{i=1}^{K}c_i\mathbf{B}_i(\mathbf{s}) \ge 0, \forall \mathbf{s} \in \twoSphere \big\}
\end{equation}
which is also a subset of the sphere manifold $\mathbb{S}^{K-1}$. The sphere, being a simple and well-studied manifold, makes the Log-Euclidean framework for ODFs computation straight-forward and efficient as seen in \equaref{eq:odf_log_map} and \equaref{eq:odf_exp_map} below:
\begin{align}
%\footnotesize
 & \forall \cc \in PS_{\!K} \subset S^{k-1}, \,  \logU(\cc) = \frac{\cc-\uu\,\mathrm{cos\Psi}}{\,\|\cc-\uu\,\mathrm{cos\Psi}\|_2}\Psi, \nonumber\\%\in \, \Tu,
 &\qquad \text{ where } \Psi = \mathrm{arcos}(\left \langle \mathbf{u}|\mathbf{c} \right \rangle)\label{eq:odf_log_map}\\
%\footnotesize
  & \forall \vc \in \Tu, \,  \expU(\vc) = \uu \, \mathrm{cos \, \Psi} + \frac{\vc}{\,\|\vc\|_2} \, \mathrm{sin} \, \Psi, %\in PS_{\,K} \subset S^{k-1},
  \text{ where } \Psi = \|\vc\|_2\label{eq:odf_exp_map}.
\end{align}
Here, $\uu$ is the uniform orientation distribution function defined as $\uu=(1,0,\ldots,0)$. We use the maps in \equaref{eq:odf_log_map} and \equaref{eq:odf_exp_map} to accurately learn ODFs and ensure their validity throughout the training process. Furthermore, the Log-Euclidean framework offers a simple geodesic estimation between two parameterized ODFs $\probODF{\cc}$ and $\probODF{\cc'}$ as follows:
\begin{equation}\label{eq:odf_log_distance}
   dist\big(\probODF{\cc}, \probODF{\cc'}\big) \, = \, \norm{\logU(\cc)-\logU(\cc')}_2.
\end{equation}

%%\subsubsection{Geometric Anisotropy}

%%Fractional anisotropy is an important DT-derived scalar that measures the degree of anisotropy of the diffusion process. In other word, it measures how non-spherical a diffusion tensor is. An equivalent measure for ODFs, the geometric anisotropy, can be defined as
%%\begin{equation}\label{eq:odf_geo_anisotropy}
%%   dist\big(\probODF{\cc}, \probODF{\uu}\big) \, = \, \norm{\logU(\cc)}_2.
%%\end{equation}
%%Similar to the FA, geometric anisotropy measure the distance of an orientation distribution function $\probODF{\cc}$ with respect to the the uniform ODF $\probODF{\uu}$.

\subsection{Adversarial Training}

\begin{figure*}[t!]
\centering
\includegraphics[width=0.75\linewidth]{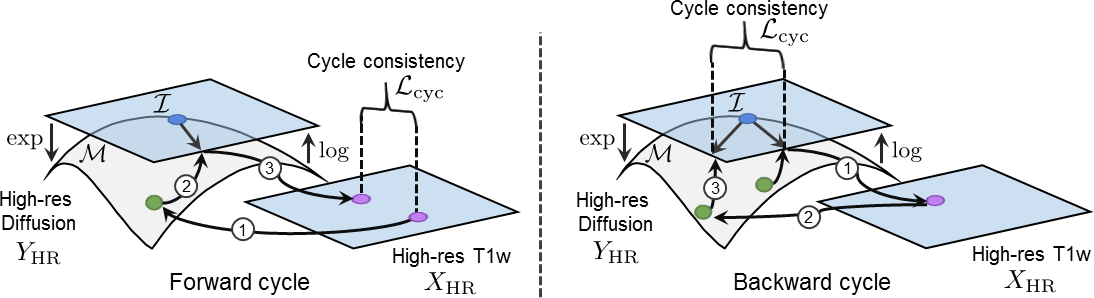}
\caption{\textbf{On the left}, our forward cycle: A T1w image is translated to a high-resolution DT on the $\spdnMan{3}$ manifold and back to the T1w domain where the cycle consistent loss is computed. \textbf{On the right}, our backward cycle: An upsampled DT on $\Tid$ is translated to the T1w domain and back to $\Tid$ where the cycle consistent loss is computed.} \label{fig:cycle_consistency}
\end{figure*}

In a standard GAN setup \citep{Goodfellow2014}, a generator network $G$ tries to generate samples so close to the true data distribution that a discriminator network $D$ is unable to distinguish between real and fake examples. Following the CycleGAN architecture, our method uses two generators and two discriminators denoted as $\genX, \genY, \disX$ and $\disY$. Here, $\genX$ takes a batch of upsampled diffusion volumes as input and tries to fool $\disX$ by generating realistic high-resolution T1w volumes. Similarly, $\genY$ takes HR T1w volumes as input and tries to generate plausible diffusion volumes in the same HR space. 

Because we only have access to LR diffusion data, the synthesized HR diffusion is downsampled using a learned residual function $\down$ prior to $\disY$'s assessment as shown in \figref{fig:architecture}. Our adversarial objectives follow the LSGAN formulation in \citep{Mao2017} and are expressed as follows:
\begin{align}\label{eq:adversarial_loss}
    &\argmin_{\disY}\,\Lgan(\genY, \disY, \XHR, \YLR) \ = \nonumber \\[-.25em]
        & \qquad \tfrac{1}{2}\,\EE_{\yy\sim\Prob_{\YLR}}\big[(\disY(\log(\yy))-1)^2\big]
         + \tfrac{1}{2}\,\EE_{\xx\sim\Prob_{\XHR}}\big[(\disY(\down(\genY(\xx))))^2\big]\\[.25em]
    &\argmin_{\disX}\,\Lgan(\genX, \disX, \YLR, \XHR) \ = \nonumber \\[-.25em]
        & \qquad \tfrac{1}{2}\,\EE_{\xx\sim\Prob_{\XHR}}\big[(\disX(\xx)-1)^2\big]
         + \tfrac{1}{2}\,\EE_{\yy\sim\Prob_{\YHR}}\big[(\disX(\genX(\!\uparrow\!\log(\yy))))^2\big]
\end{align}
where $\uparrow$ represents trilinear upsampling and $\log$ is the logarithm map defined in either \equaref{eq:log_of_tensor} or \equaref{eq:odf_log_map}. It should be noted that both the upsampling of the real data and the discriminator evaluation of the generated diffusion information are performed in the Log-Euclidean domain to account for the underlying data manifold.

\subsection{Cycle-consistency Loss}

The adversarial losses alone are not sufficient to drive the generation of HR diffusion data. Indeed, $\disY$ only evaluates downsampled data and, thus, cannot help $\genY$ improving beyond a certain precision level. Therefore, the cycle-consistency loss denoted in \equaref{eq:cycle_loss} not only helps ensuring the structural coherence of the synthesized images across modalities, but also provides important high-resolution gradients to train $\genY$. 

Our cycle-consistency loss is threefold: 1) the error between the original HR structural volume $\xx$ and the reconstructed volume $\genX(\genY(\xx))$, 2) the error between the upsampled diffusion $\!\uparrow\!\!\log(\yy)$ and its HR reconstruction $\genY\big(\genX(\!\uparrow\!\log(\yy))\big)$, and 3) the error between the original LR diffusion $\yy$ and the downsampled recovered volume $\down\big(\genY\big(\genX(\uparrow\!\log(\yy))\big)\big)$. Combining these in a single loss gives
\begin{align}\label{eq:cycle_loss}
    & \Lcyc(\genY, \genX) \ = \ \lambda_{\mr{cyc}_{X}} \underbrace{\EE_{\xx \sim \Prob_{\XHR}} \big[\norm{\genX(\genY(\xx)) \, - \, \xx}_1\big]}_\text{Forward Cycle HR} \nonumber\\%[-0.1em]    
   & \qquad \quad \ + \ \tfrac{1}{2}\lambda_{\mr{cyc}_{Y}} \underbrace{\EE_{\yy \sim \Prob_{\YLR}} 
    \big[\norm{\genY\big(\genX(\uparrow\!\log(\yy))\big) \, - \, \!\uparrow\!\log(\yy)}_1\big]}_\text{Backward Cycle HR} \nonumber\\%[-0.1em]
   & \qquad \qquad \quad \ + \ \tfrac{1}{2}\lambda_{\mr{cyc}_{Y}} \underbrace{\EE_{\yy \sim \Prob_{\YLR}} 
    \big[\norm{\down\big(\genY\big(\genX(\uparrow\!\log(\yy))\big)\big) \, - \, \log(\yy)}_1\big]}_\text{Backward Cycle LR}
\end{align}

We employ the $\ell_1$ norm in \equaref{eq:cycle_loss} to measure both the forward and backward cycle reconstruction errors, as it is less sensitive to large errors than the $\ell_2$ norm \citep{Zhao2016}. Again, the log map is used to project the generated and the real manifold-valued data onto a tangent plane before computing the cycle-consistency loss. Furthermore, two parameters $\lambda_{\mr{cyc}_{X}}$ and $\lambda_{\mr{cyc}_{Y}}$ control the contribution of both cycles and have been empirically tuned. Full cycles, including the manifold mappings and the loss computation in the tangent plane, are illustrated in \figref{fig:cycle_consistency}.

\subsection{Image Prior Regularization}

By using a cycle-consistency loss in both directions, the CycleGAN model is able to learn a bijective mapping between two domains using unpaired examples \citep{Zhu2017}. However, for many cross-domain translation problems, the solution space is extremely large and the model does not necessarily converge to a solution that satisfies important domain-specific properties \citep{Lu_Zhou_Song_Ren_Yu_2019}. This is problematic, especially in the case of medical images synthesis where the generated images must not only be realistic from the discriminator's point of view, but also be faithful to expected results of the downstream tasks and known anatomical properties. Thus, to ensure the model's convergence towards plausible solutions, we introduce a prior loss as follows:
\begin{align}\label{eq:prior_loss}
    & \Lprior(\genY, \genX) \ = \ \lambda_{\mr{prior}_{X}} \EE_{\xx \sim \Prob_{\XHR}, \yy \sim \Prob_{\YLR}} \big[\norm{\genY(\xx_i) \, - \, \uparrow\!\log(\yy_i)}_1\big] \nonumber \\[.1em]
    & \qquad \quad \ + \ \lambda_{\mr{prior}_{Y}} \EE_{\xx \sim \Prob_{\XHR}, \yy \sim \Prob_{\YLR}} \big[\norm{\genX(\uparrow\!\log(\yy_i)) \, - \, \xx_i}_1\big]
\end{align}
where $\xx_i$ and $\yy_i$ are paired volumes taken from a limited number of subjects. With this loss, the super-resolved diffusion stays close to the real upsampled diffusion while integrating high-frequency elements from the HR structural images.

\subsection{Diffusion Anisotropy Weighted Loss}\label{sec:fa-weighted-loss}

\begin{figure}[t!]
\centering
\includegraphics[width=.99\linewidth]{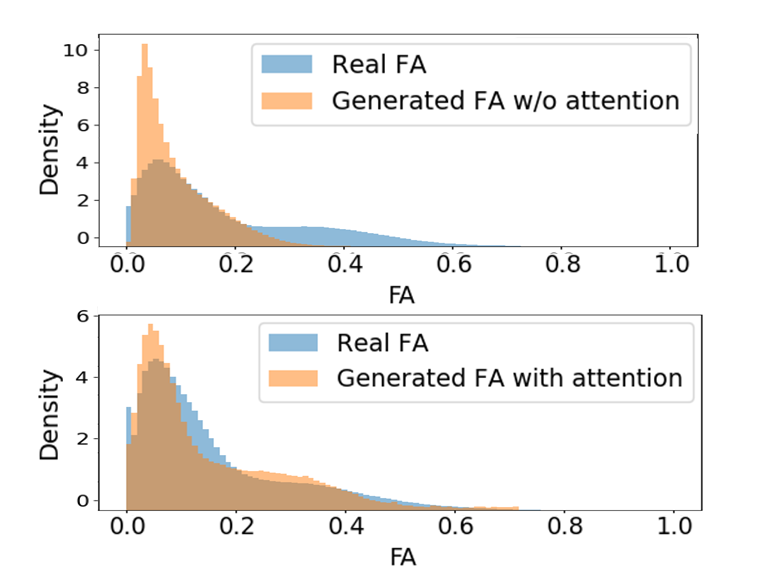}
\caption{Comparison of the real FA (blue) and the generated FA (orange) distribution density of the same subject with and without diffusion anisotropy weighted loss.} \label{fig:fa_hist_attention}
\end{figure}

Voxels expressing fiber tracts information typically have higher FA values than those representing other tissues like grey matter (GM) or cerebrospinal fluid (CSF). Therefore, we would like our diffusion synthesis method to be particularly accurate in regions with higher fractional anisotropy. However, as seen in \figref{fig:fa_hist_attention}, voxels with high FA are underrepresented compared to those with lower values. Consequently, this imbalance problem drives the network's generation towards diffusion with FA close to the mean. To alleviate this issue, we weight the diffusion error in $\Lcyc$ and $\Lprior$ by the FA/GFA of the target volume at every voxel. The benefit of such mechanism can be observed on the density plots at the bottom of \figref{fig:fa_hist_attention} which clearly exhibit a more faithful FA distribution when using the proposed diffusion anisotropy weighting scheme.

\subsection{Full Objective}

Combining all loss terms, our full objective function is given by
\begin{align}
   \Loss \ & = \ -\, \Lgan(\genX, \disX, \YLR, \XHR) \, - \Lgan(\genY, \disY, \XHR, \YLR)\nonumber\\
    & \qquad\qquad + \, \Lcyc(\genY, \genX) \, + \, \Lprior(\genY, \genX)\label{eq:full_objective}
\end{align}
As in standard adversarial learning approaches, we train the generators and discriminators concurrently by solving a mini-max problem:
\begin{equation}\label{eq:full_min_max}
\begin{split}
   \genX^*, \genY^* \, = \, \arg\argmin_{\genX, \genY} \argmax_{\disX, \disY}\Loss(\genX, \genY, \disX, \disY)
 \end{split}
\end{equation}

Hence, \equaref{eq:full_objective} combines the error feedback from both the HR structural images and the LR diffusion in an adversarial and voxel-wise manner.

\section{Results}\label{sec:experiments}

\subsection{Data and Pre-Processing}

We employ the T1w and diffusion MRI data of 1,065 subjects from the HCP1200 release of the Human Connectome Project \citep{VanEssen2013} to evaluate our manifold-aware CycleGAN. The T1w (0.7mm$^3$ voxels, FOV=224mm, matrix=320, 256 sagittal slices in a single slab) and diffusion (1.25mm$^3$ voxels, sequence=Spin-echo EPI, repetition time (TR)=5520 ms, echo time (TE)=89.5ms) data were acquired with a Siemens Skyra 3T scanner \citep{Sotiropoulos2013} and minimally processed following \citep{Glasser2013}.

Diffusion tensors were fitted using the DSI Studio toolbox software \citep{Jiang2005} and the dODFs estimated using the constant solid angle (CSA) method \citep{Aganj2010} from the DIPY library \citep{Garyfallidis2014}. Diffusion ODFs were then re-parameterized following \secref{sec:odfs_reparameterization} and further estimated using 4th order spherical harmonics \citep{Descoteaux2007}. Both the DT and the ODF volumes were transformed to the Log-Euclidean domain using their respective Riemannian framework described in \secref{section:Method}. Once in the Log-Euclidean domain, these volumes were upsampled to the T1w spatial resolution using trilinear interpolation and aligned to their corresponding HR T1w images. In experiments, we consider these upsampled and aligned diffusion volumes as the ``ground truth'' diffusion. The T1w images have been rescaled to the [0,1] range by min-max normalization. Finally, both the structural and the diffusion volumes, in high and low-resolution, were decomposed in overlapping patches of $32^3$ and $18^3$ voxels respectively. Volumes are processed patch-wise by our model for two important reasons: 1) limiting the memory required by the model to compute network activations and outputs, and 2) increasing the amount of training examples.

\begin{figure*}[t!]
\centering
\includegraphics[width=.95\linewidth]{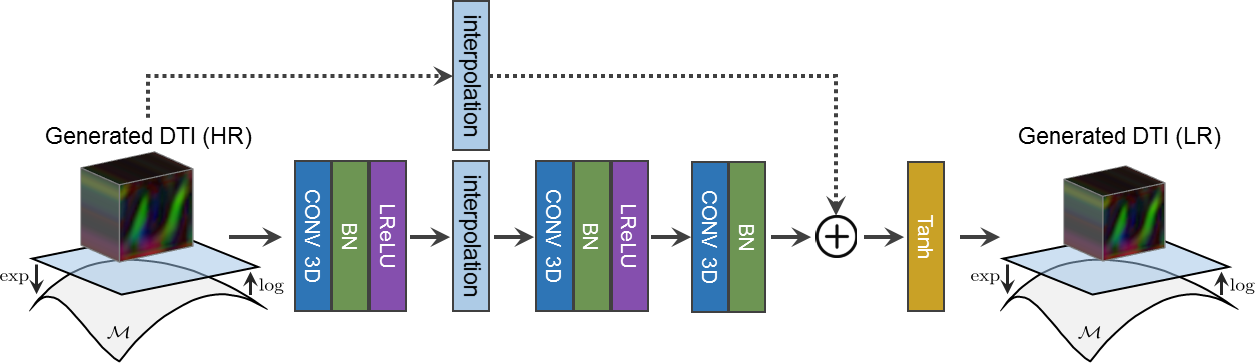}
\caption{The architecture of our residual downsampling function $\mathcal{F}$ used to reduce the resolution of our synthesized HR diffusion before $\disY$ assessment.} \label{fig:downsampling}
\end{figure*}

\subsection{Implementation Details}

Our two generator networks, $\genX$ and $\genY$, follow the U-Net implementation in \citep{Cicek2016} where the last activation layer has been replaced to fit the different output data scale. In all the experiments, we used a sigmoid in $\genX$ to generate T1w images within the [0,1] range. For the generation of diffusion, the last activation function of $\genY$ varies depending on the generated diffusion model. For the generation of DT, we used a hard hyperbolic tangent function and for the generation of ODF, a tanh activation.

Moreover, we set the amount of input and output channels of our networks according to our input and output data shape as described in \tabref{tab:architecture_details}. The DT inputs are of shape $9 \times 32 \times 32 \times 32$, where the 9 channels represent the flattened $3 \times 3$ diffusion tensors at every voxel. The ODF inputs are of shape $C \times 32 \times 32 \times 32$ where C depends on the order of the spherical harmonic basis that is used to represent them. Due to limitations on computational resources, we estimated the diffusion ODFs using 4th order spherical harmonics, yielding a total of $C=15$ coefficients.

Our discriminator networks $\disX$ and $\disY$ follow the SRGAN discriminator architecture in \citep{Ledig} where we replaced the 2D convolutions by 3D ones. Furthermore, we reduced the number of feature maps in convolution layers to 32, 64, 128 and 256 as suggested in \citep{sanchez2018brain} for volumetric data. In all scenarios, $\disX$ assesses T1w volumes of shape $1 \times 32 \times 32 \times 32$ while $\disY$ takes as inputs volumes of shape $9 \times 18 \times 18 \times 18$ for DT and $15 \times 18 \times 18 \times 18$ for ODFs.

\subsection{Training Setup}

To train our network, we randomly selected 70\% (746 subjects) of the 1,065 subjects for training, 20\% (213 subjects) for validation and 10\% (106 subjects) for testing. From the 745 training subjects, we kept aside between 0 and 75 subjects as paired priors in \equaref{eq:prior_loss}. To form our training, validation and test sets, we randomly selected 50,000 unpaired HR T1w and LR diffusion patches from remaining training subjects, 10,000 patches from validation subjects and 5,000 patches from test subjects. The same number of patches (50,000) was selected from our subjects kept as paired priors, i.e. aligned HR T1w and upsampled diffusion, to form a paired training set. Although we kept the number of patches in the paired training set constant, we experimentally adjusted the number of subjects used to randomly extract them. To do so, we trained our network with an increasing number of paired subjects, as reported in \tabref{tab:paired_subjects_results}, and kept the best setup for our following experiments.

\begin{table*}[t!]
\centering
\caption{Implementation details of our generator networks $\genX$ and $\genY$ and discriminator networks $\disX$ and $\disY$.}
\label{tab:architecture_details}
\begin{tabular}[t]{lC{3.5cm}C{3.5cm}C{3cm}}
\toprule
\textbf{Network} & \textbf{Input Shape} & \textbf{Output Shape} & \textbf{Last Activation} \\
\midrule
\midrule
$\genY$ (T1w $\mapsto$ DT) & $\phm1\times 32\times 32\times 32$  & $\phm9\times 32\times 32\times 32$  &  Hardtanh\\
$\genX$ (DT $\mapsto$ T1w) & $\phm9\times 32\times 32\times 32$  & $\phm1\times 32\times 32\times 32$  &  Sigmoid\\
$\genY$ (T1w $\mapsto$ ODF) & $\phm1\times 32\times 32\times 32$  & $15\times 32\times 32\times 32$ & Tanh\\
$\genX$ (ODF $\mapsto$ T1w) & $15\times 32\times 32\times 32$ & $\phm1\times 32\times 32\times 32$  &  Sigmoid\\
\midrule
\midrule
$\disX$ (T1w) & $\phm1\times 32\times 32\times 32$  & 1 & Linear\\
$\disY$ (DT) & $\phm9\times 18\times 18\times 18$  & 1 & Linear\\
$\disY$ (ODFs) & $15\times 18\times 18\times 18$  & 1 & Linear\\
\bottomrule
\end{tabular}
\end{table*}

We trained the networks using an Adam optimizer \citep{Kingma2015} with a learning rate of $10^{-4}$ and beta1, beta2 values of 0.5 and 0.999. Hyper-parameters $\lambda_{\mr{prior}_{X}}$, $\lambda_{\mr{prior}_{Y}}$, $\lambda_{\mr{cyc}_{X}}$ and $\lambda_{\mr{cyc}_{Y}}$ were experimentally set to 10, 0.5, 5 and 0.25. Furthermore, we used a reduce on plateau learning rate scheduler with a patience of 10 epochs and a factor of 10. Batches of 8 patches were used and the models were trained for 35 epochs ($\sim 220$k steps) on an NVIDIA TITAN XP GPU with 12 GB of VRAM. All experiments were repeated three times with a different initialization seed.

\subsection{Baselines}

As mentioned before, deep learning models for the synthesis of manifold-valued data are just starting to emerge. Consequently, the number of baselines requiring minimal adaptation for the evaluation of our model is limited. Nevertheless, we compare our model to the three approaches described below.

\paragraph{Manifold-Aware WGAN}
Our first baseline is an adaptation of the manifold-aware WGAN presented in \citep{Huang2019} for the conditional generation of diffusion from structural T1w images. This method denoted as ``MA-WGAN'' in our results, can generate plausible manifold-valued images by incorporating the Log-Euclidean maps within the network and therefore provides a natural point of comparison for our method. For this baseline, we use the same generator $\genY$ and discriminator $\disY$ as in our proposed model. The manifold mapping used in the network is changed according to the generated diffusion scheme following \secref{section:Method}.

\paragraph{Manifold-Aware U-Net}

We also compare our method to a supervised U-Net model. For this baseline, we use the same generator as in our own architecture, i.e. $\genY$~\citep{Cicek2016}, but train it in a supervised manner with paired HR T1w and upsampled diffusion volumes in the Log-Euclidean domain. Similar to the \emph{MA-WGAN} baseline, we change the manifold mapping of the network according to the generated diffusion reconstruction scheme (DT or ODF). This baseline, denoted as ``MA-U-Net'' in results, helps us measure the effect of our adversarial and cycle-consistent losses.

\paragraph{U-Net}

Our last baseline is a standard supervised U-Net~\citep{Cicek2016} trained with paired HR T1w images and upsampled diffusion without manifold-awareness. With this method, we aim at measuring the performance gain of our method induced by both the manifold-mapping and the use of additional unpaired samples. In addition, we validate that manifold-awareness is necessary to synthesize realistic samples strictly lying on the data manifold.

\subsection{DT and ODF Synthesis}

We first test our network and baselines for the task of DT and diffusion ODF synthesis. In this setup, we train our network with unpaired HR T1w patches and LR diffusion patches in the Log-Euclidean domain. Moreover, we use a paired training set of 50\,000 patches from 50 randomly chosen subjects. The paired training set is used as prior for our method and as the training set for our baselines. Hence, both our method and baselines are trained with the same amount of paired information.       

\paragraph{Evaluation Metrics}\label{sec:evaluation_metric}

Three metrics were considered to quantitatively evaluate the generated diffusion. First, we use the cosine similarity to compare the principal fiber orientation of every synthesized tensor and ODF to their expected real orientation:
\begin{equation}\label{eq:cosine_similarity}
\begin{split}
   \mathrm{similarity}(\matr{a},\matr{b}) \, = \, \abs*{\frac{\matr{a} \cdot \matr{b}}{\norm{\matr{a}}\norm{\matr{b}}}}
 \end{split}
\end{equation}
To retrieve the main orientation of the ODF, we first calculate the spherical coordinates at which the value of the ODF is maximum using a discretized sphere of 724 vertices. We then convert these spherical coordinates to Euclidean coordinates to obtain their principal orientation vector. For DT, the main orientation is given by the eigenvector associated with the largest eigenvalue of the DT as described in \secref{sec:dti_riemannian_framework}. 

\begin{figure}[hbt!]
\centering
\includegraphics[width=0.99\linewidth]{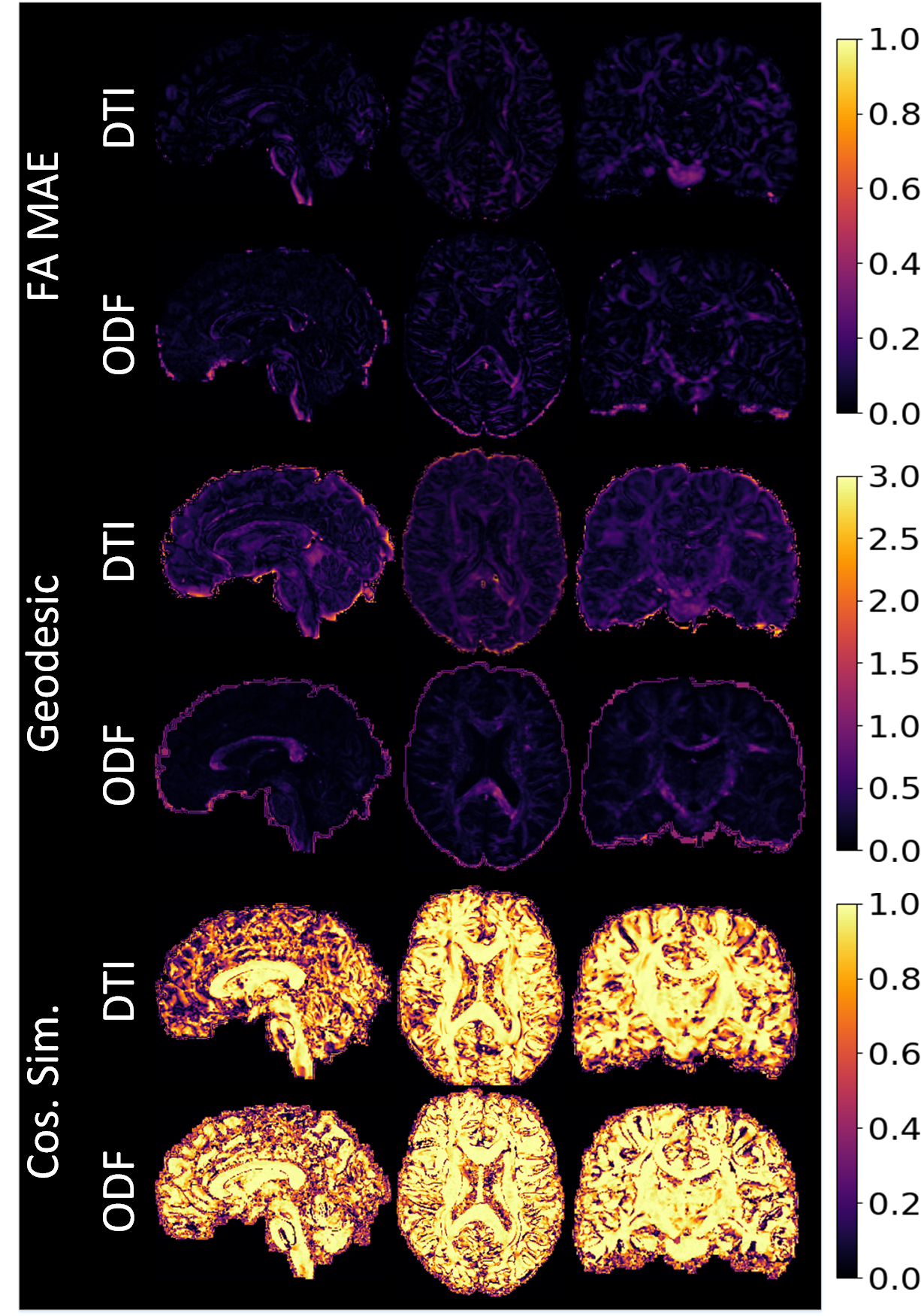}
\caption{Metrics on the sagittal, coronal and axial slices between the generated HR diffusion of a random test subject and its interpolated ground-truth. (\textbf{Top row}) FA, (\textbf{middle row}) geodesic distance and (\textbf{bottom row}) cosine similarity.} \label{fig:metric_map}
\end{figure}

In addition, we compare the generated and real diffusion with the mean square error (MSE) between their FA/GFA. The FA and GFA are computed following \equaref{eq:fractional_anisotropy} and \equaref{eq:generalized_fractional_anisotropy}. This help evaluating the shape of the synthesized DT/ODF independently of their orientation.

Finally, we measure the mean geodesic distance using \equaref{eq:spd_log_distance} for DT and \equaref{eq:odf_log_distance} for ODF between our generated data and the real diffusion. This latter metric encode both the orientation and the shape error in a single measure.

\subsection{Tractography}

To further assess the integrity of the synthesized diffusion volumes by the proposed method, we performed whole-brain tractography on both the real and the generated data, and segmented the resulting tractograms into bundles. We then posed the tractograms generated on real data as ground truth and extracted quantitative measures from whole-brain tractograms. Likewise, we segmented bundles to measure how much is tractography impacted. In the following subsections, we describe each step of the analysis.

\begin{figure*}[t!]
\centering
\includegraphics[width=0.99\linewidth]{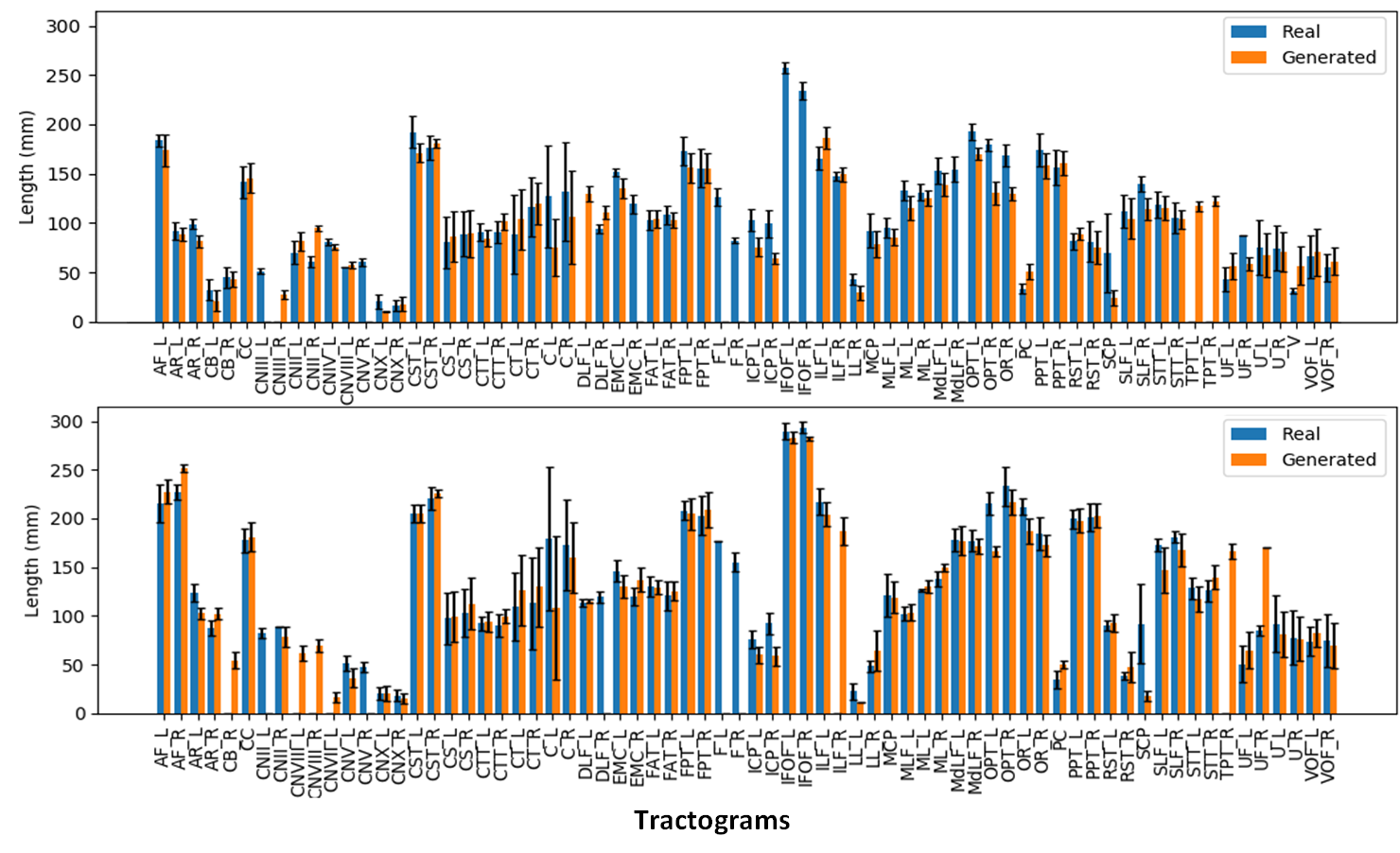}
\caption{Mean streamline length (in mm) for the segmented bundles from DT tractography (\textbf{top row}) and ODF tractography (\textbf{bottom row}).} \label{fig:results_streamline_len}
\end{figure*}

\paragraph*{Streamlines Generation and Bundling}
    
Tracking was performed using the EuDX algorithm~\citep{Garyfallidis2013} with a step-size of 0.5\,mm. A maximum angle of 60 degrees was used between steps, using the principal direction of the diffusion tensor and maxima of ODF. Maxima were extracted from the ODF using scilpy \footnote{\url{https://github.com/scilus/scilpy}}. Seeding was done at 2 seeds per voxel on the whole white-matter mask, which was computed from the ground-truth T1w image using Dipy~\citep{Garyfallidis2014}. Streamlines with a length below 10\,mm or above 300\,mm were discarded. Whole brain tractograms were then segmented using RecobundlesX~\citep{Garyfallidis2018}, using 80 bundles from ~\citet{Yeh2018} as reference. To allow for a more robust comparison, and because initial streamline points placement depends on randomness which may have an impact on the reconstructed streamlines, tracking was performed five times with different random seeds on each volume.

\paragraph*{Streamlines Assessment}

\begin{figure*}[hbt!]
\centering
\includegraphics[width=0.95\textwidth]{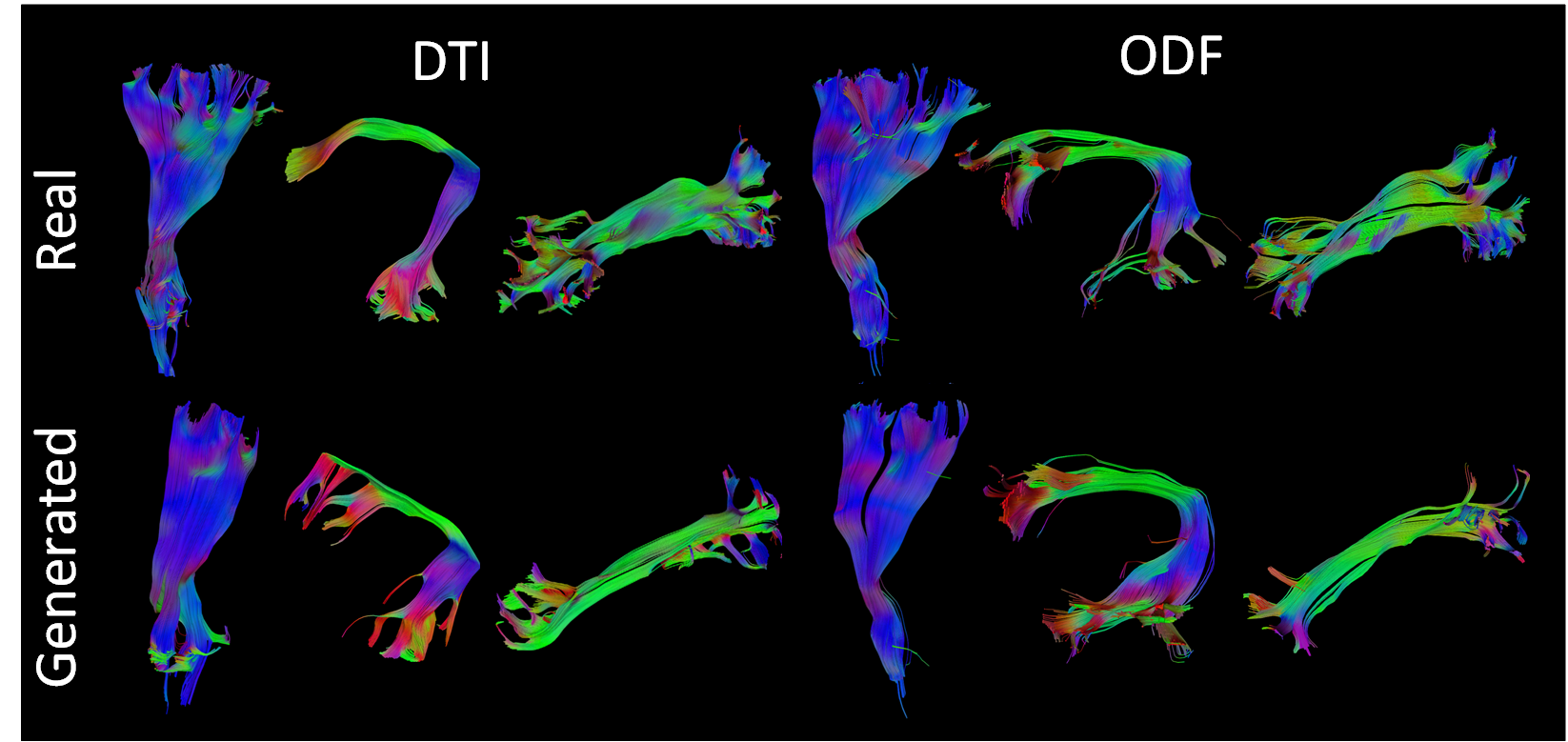}
\caption{Visualization of three segmented bundles from real and synthesized tractograms. (\textbf{On the left}), left Corticospinal Tract (CST), Arcuate Fasciculus (AF) and Inferior Longitudinal Fasciculus (ILF) segmented from real and generated DT data. (\textbf{On the right}), left CST, AF and ILF segmented from real and synthesized ODF data.} \label{fig:dti_visualization}
\end{figure*}

The similarity between the streamlines reconstructed from the real and generated diffusion volumes is assessed using three measures: streamline length, bundle volume and bundle shape. First, we compare the streamline length (in mm) between all real and synthesized whole-brain tractograms, as well as each segmented bundle. We also compare the volume occupied by the whole-brain tractograms and bundles in a voxel-wise matter. Furthermore, the shape similarity of reconstructed tractograms is measured by their voxel-wise Dice, overlap (OL) and overreach (OR). The OL, defined as

\begin{equation}\label{eq:overlap}
\begin{split}
   \mathrm{OL} \, = \, \frac{|B \cap A|}{|A|},
 \end{split}
\end{equation}
where $A$, $B$ are binary bundle masks, quantifies how much the volume of bundle $A$ is reconstructed by bundle $B$. The OR, expressed as
\begin{equation}\label{eq:overreach}
\begin{split}
   \mathrm{OR} \, = \, \frac{|B \cup A| - |B \cap A|}{|A|},
 \end{split}
\end{equation}

evaluates how much of bundle $B$ goes over bundle $A$. Segmented bundles were merged to allow for pairwise comparison between real and generated data. Merged bundles with fewer than 100 streamlines were discarded from the analysis. To evaluate the overall reconstruction quality, we also report Dice, OL and OR between the reconstructed tractograms and the atlas used for bundle segmentation.

Figure \ref{fig:dti_visualization} presents some of the reconstructed bundles from real and synthesized data. 

\subsection{Diffusion Synthesis Analysis}

We report in \tabref{tab:paired_subjects_results} the performance of our model with an increasing number of paired subjects used as prior in \equaref{eq:prior_loss}. We observe that for both DT and ODF, the mean cosine similarity increases with the number of subjects and reaches its peak when extracting the paired patches from 50 subjects. As we keep the number of extracted patches constant (50,000 patches), 50 subjects represent an average of 1,000 patches per subject which seems a good trade-off between sampling diversity and sparsity. In this setup, our model yields a mean cosine similarity of 0.8648 $\pm\,\num{5.5e-3}$ and 0.8846 $\pm\,\num{4.4e-3}$ in voxels with $\mr{FA} \ge 0.2$ for DT and ODF respectively. In voxels with $\mr{FA} \ge 0.5$, a mean cosine similarity of 0.9167 $\pm\,\num{3.2e-3}$ is reached for DT and 0.9425 $\pm\,\num{1.4e-4}$ for ODF. As reported in \tabref{tab:metrics_results}, this corresponds to FA MSE values of 0.0089 $\pm\,\num{1.5e-3}$ and 0.0159 $\pm\,\num{1.4e-4}$ for DT and GFA MSE values of 0.0229 $\pm\,\num{2.8e-4}$ and 0.0614 $\pm\,\num{1.3e-3}$ for ODF.

\begin{table}[t]
\caption{The cosine similarity between the principal orientation obtained by our method with different amount paired subjects used as prior in \equaref{eq:prior_loss}. Since the metrics are more relevant in regions that typically encode fibers information, we report them at increasing FA thresholds of 0.2 and 0.5.}\label{tab:paired_subjects_results}
\centering
\begin{small}
\renewcommand{\arraystretch}{1.1}
\begin{tabular}[t]{ccccc}
%{C{1.7cm}C{1.4cm}C{1.4cm}C{1.4cm}C{1.4cm}C{1.4cm}C{1.4cm}}
\toprule
\multirow{2}{*}[-.75em]{\shortstack{\textbf{Paired} \\[2pt] \textbf{Subjects}}} & \multicolumn{2}{c}{\textbf{Cosine Sim (DT)}} & \multicolumn{2}{c}{\textbf{Cosine Sim (ODF)}}\\
\cmidrule(lr){2-3}
\cmidrule(lr){4-5}
 & FA\,$\geq$\,0.2 & FA\,$\geq$\,0.5 &  GFA\,$\geq$\,0.2 & GFA\,$\geq$\,0.5 \\
\midrule
0 & 0.7315 & 0.7955 & 0.7029 & 0.7380\\
10 & 0.7795 & 0.8599 & 0.8664 & 0.9058\\
25 & 0.8152 & 0.8863 & 0.8745 & 0.9081\\
\textbf{50} & \textbf{0.8648} & \textbf{0.9167} & \textbf{0.8846} & \textbf{0.9425}\\
75 & 0.8317 & 0.8993 & 0.8674 & 0.9051\\
\bottomrule
\end{tabular}
\end{small}
\end{table}

As a point of comparison, we give in \tabref{tab:metrics_results} the performance of our baselines when they are trained with the same 50 paired subjects. As can be seen, the proposed model obtains the highest performance for all metrics when trained on DT images, as well as better geodesic and fiber orientation estimation than baselines when trained on ODF.

Compared to the fully-supervised U-Net, which does not enforce manifold consistency on the output, our model improves FA MSE by 23.14\%, mean geodesic distance by 81.11\% and mean cosine similarity by 4.23\% for regions with $\mr{FA} \ge 0.5$ for DT. This shows the benefit of imposing manifold-awareness constraints on the network's output. Without these constraints, U-Net generates an average of 3,843.5 non-SPD tensors and 1,559.7 non-PDF ODF as noted in \tabref{tab:non_manifold_count}. Our model also provides statistically better performance, for both DT and ODF data, on the mean cosine similarity and mean geodesic distance compared to MA-U-Net (paired t-test p $<$ 0.05), that does not include cycle-consistency and is only trained with paired data. 

\begin{table}[t]
\caption{The amount of non-SPD tensors and non-PDF dODFs generated by the compared methods for five randomly selected test subjects. The manifold-awareness ensures that the generated diffusion schemes lie on their respective data manifold.}\label{tab:non_manifold_count}
\centering
\begin{footnotesize}
\renewcommand{\arraystretch}{1.1}
\begin{tabular}[t]{lC{1.5cm}C{2.0cm}C{2.0cm}}
\toprule
\textbf{Method} & \textbf{Manifold-Awareness} & \textbf{Non-SPD Tensors} & \textbf{Non-PDF ODF}\\
\midrule
\midrule
U-Net & \xmark & $3843.5 \pm 481.22$ & $1559.7 \pm 122.3$\\
\textbf{MA-U-Net} & \cmark & \textbf{0} & \textbf{0}\\
\textbf{MA-WGAN} & \cmark & \textbf{0} & \textbf{0}\\
\textbf{Ours} & \cmark & \textbf{0} & \textbf{0}\\
\bottomrule
\end{tabular}
\end{footnotesize}
\end{table}

Improvements are particularly important for voxels with $\mr{FA} \ge 0.5$, where our method obtains a 3.43\% higher mean cosine similarity, 11.19\% lower geodesic and 13.58\% better FA MSE for DT and 4.18\% higher mean cosine similarity for ODF. This demonstrates the impact of our anisotropy-weighted loss described in \secref{sec:fa-weighted-loss}, which gives more importance to voxels with higher anisotropy values. 

\begin{figure}[hbt!]
\centering
\includegraphics[width=.95\linewidth]{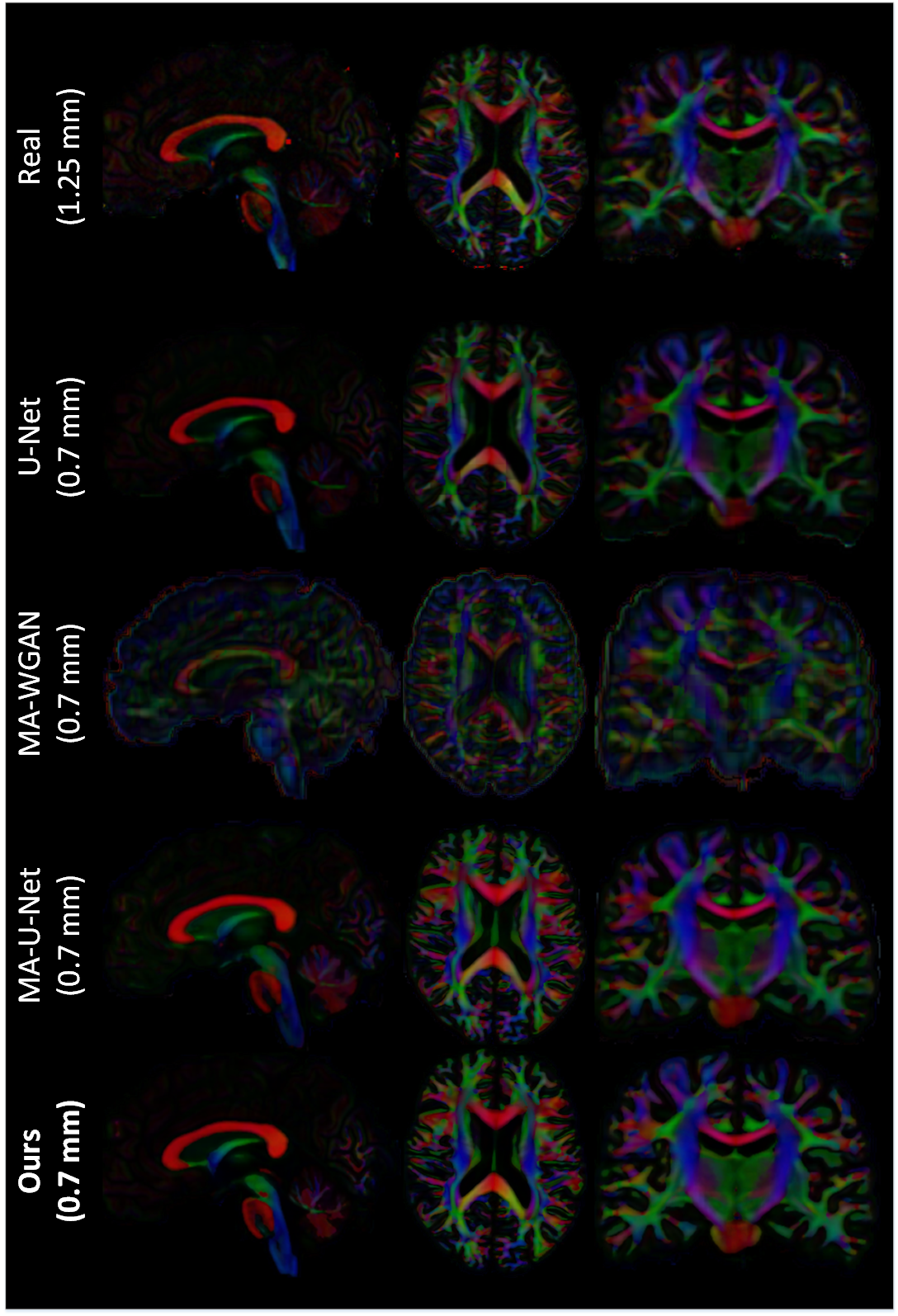}
\caption{Comparison of the color FA of the generated diffusion of compared methods on a sagittal, axial and coronal slice of a random test subject.} \label{fig:color_FA_baseline_comparison}
\end{figure}

\begin{figure*}[t!]
\centering
\includegraphics[width=.99\linewidth]{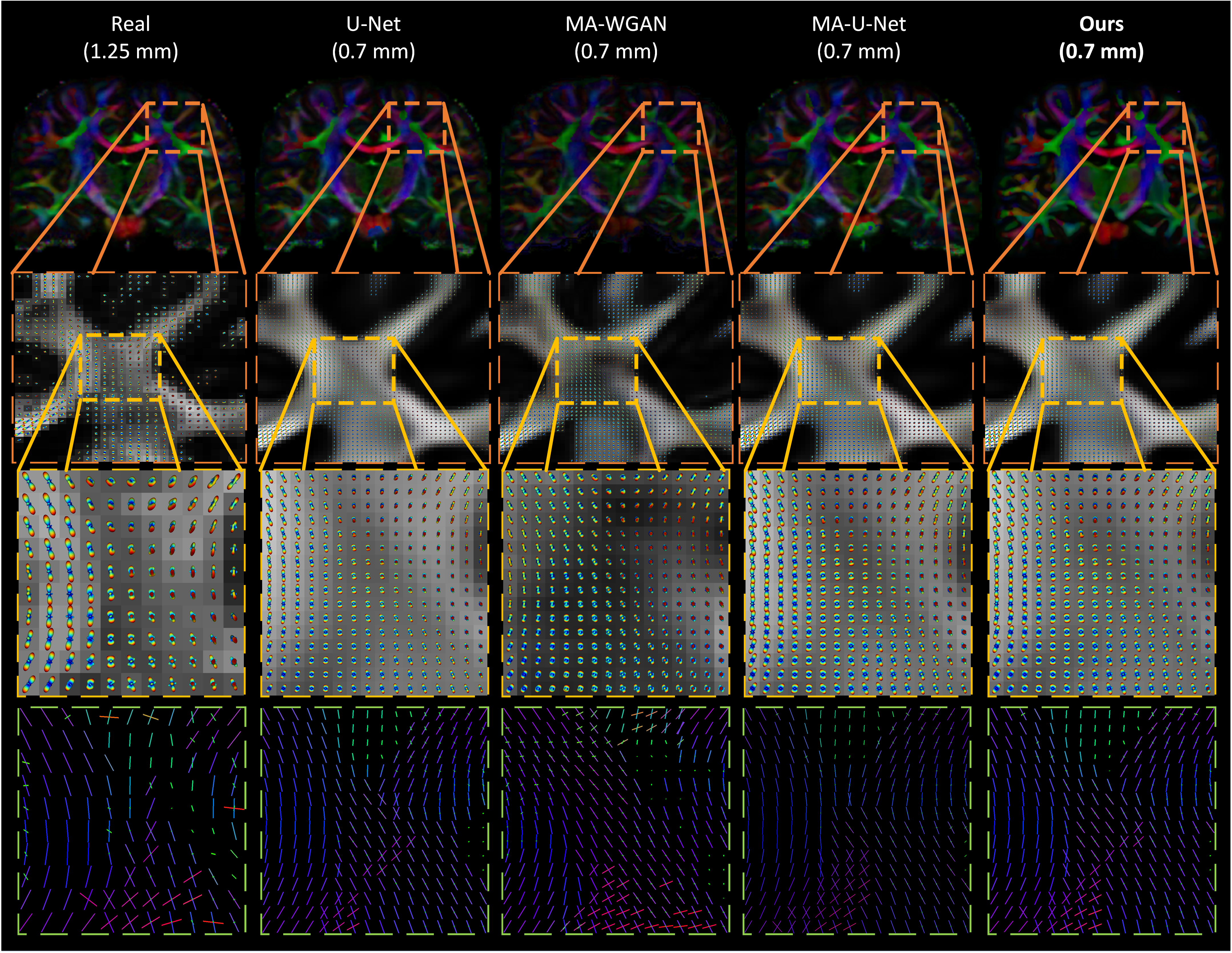}
\caption{Qualitative comparison of the generated ODF by the compared methods. \textbf{First row}, the color encoded GFA of a coronal slice of a test subject. \textbf{Second row}, the generated ODF and GFA in a region with crossing fibers. \textbf{Third row}, a close-up look of the generated ODF and GFA. \textbf{Fourth row}, a close-up look at the ODF peaks.} \label{fig:odf_crossing}
\end{figure*}

%ODF
% Improvement over U-Net (%)
%  -4.09 | -10.01 | 4.96 | 0.97 | 1.65 | 4.20
% Improvement over MA-U-Net (%)
%  -4.56 | -9.14 | 3.98 | 0.17 | 1.56 | 4.18

\begin{table*}[t]
\caption{The fractional anisotropy mean squared error (FA MSE), geodesic, and cosine similarity between the principal orientations obtained by the different compared methods when trained with 50 paired subjects. Since the metrics are more relevant in regions that typically encode fibers information, we report them at increasing FA thresholds of 0.2 and 0.5.}\label{tab:metrics_results}
\centering
\renewcommand{\arraystretch}{1.1}
\setlength{\tabcolsep}{4.5pt}
\begin{tabular}[t]{clcccccc}%{C{2.2cm}C{1.4cm}C{1.4cm}C{1.4cm}C{1.4cm}C{1.4cm}C{1.4cm}}
\toprule
& \multirow{2}{*}[-1em]{\textbf{Method}}
& \multicolumn{2}{c}{\textbf{\FA\,MSE}} &   \multicolumn{2}{c}{\textbf{Geodesic}} & \multicolumn{2}{c}{\textbf{Cosine Similarity}}\\
\cmidrule(lr){3-4}\cmidrule(lr){5-6}\cmidrule(lr){7-8}
 & & \FA\,$\geq$\,0.2 & \FA\,$\geq$\,0.5 & \FA\,$\geq$\,0.2 & \FA\,$\geq$\,0.5 & \FA\,$\geq$\,0.2 & \FA\,$\geq$\,0.5 \\
\midrule
\multirow{4}{*}{DT~} & U-Net & 0.0100 & 0.0207 & 2.2559 & 2.5457 & 0.8266 & 0.8795\\
& MA-WGAN & 0.0294 & 0.0781 & 0.7791 & 1.046 & 0.5724 & 0.6246\\
& MA-U-Net & \textbf{0.0088} & 0.0184 & 0.3947 & 0.5415 & 0.8288 & 0.8863\\
& \textbf{Ours} & 0.0089 & \textbf{0.0159} & \textbf{0.3531} & \textbf{0.4809} & \textbf{0.8648} & \textbf{0.9167}\\
\midrule
\multirow{4}{*}{ODF~} & U-Net & 0.0220 & \textbf{0.0559} & 0.3639 & 0.6299 & 0.8702 & 0.9045\\
& MA-WGAN & 0.0443 & 0.1173 & 0.5351 & 0.9121 & 0.6823 & 0.7278\\
& MA-U-Net & \textbf{0.0219} & 0.0563 & 0.3611 & 0.6249 & 0.8710 & 0.9047\\
& \textbf{Ours} & 0.0229 & 0.06145 & \textbf{0.3467} & \textbf{0.6238} & \textbf{0.8846} & \textbf{0.9425}\\
\bottomrule
\multicolumn{8}{l}{\emph{* : FA is used for DT and GFA for ODF}.}
\end{tabular}
\end{table*}

Our results can be further appreciated in \figref{fig:metric_map}, where we report the metrics yielded by our method on a sagittal, axial and coronal slice of a randomly chosen test subject. We see that our network is able to recover most of the fibers orientation, especially in regions with typically higher FA/GFA like the corpus callosum. Moreover, the estimated FA/GFA is generally faithful to the real data except in the corticospinal tract where the error is higher.

\begin{figure*}[hbt!]
    \centering
    \includegraphics[width=.95\textwidth]{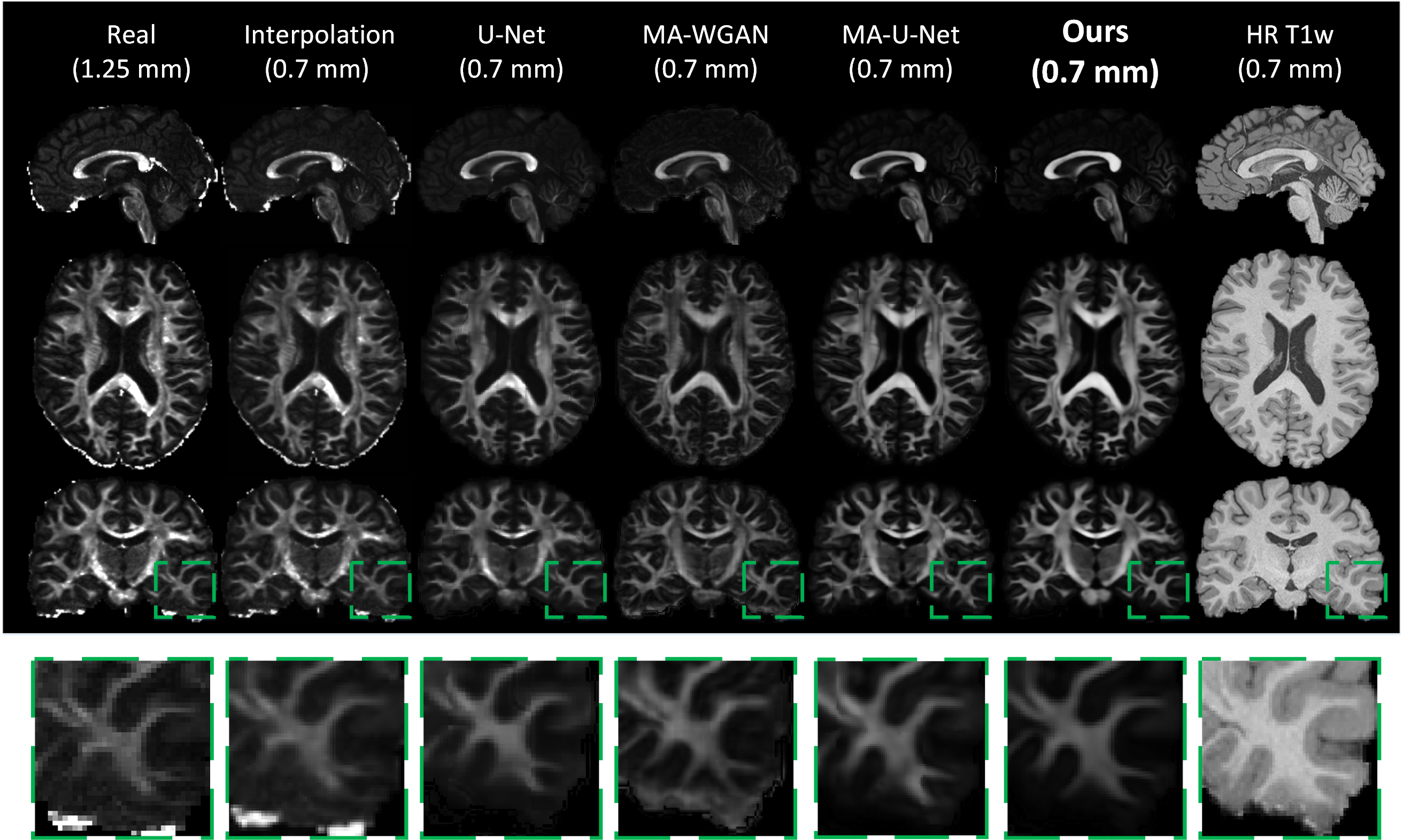}
    \caption{GFA of the real low-resolution diffusion (\textbf{first column}), the real upsampled diffusion (\textbf{second column}) and the generated diffusion of the compared methods. One can notice the ability of our network to recover fine anatomical details that are clearly visible in the HR T1w image but not in the original diffusion.} \label{fig:gfa_comparison}
    \end{figure*}

Finally, as a qualitative evaluation, we compare in \figref{fig:color_FA_baseline_comparison} the generated color encoded FA of compared methods. From this figure, we can see how our cycle-consistent and prior losses help our model converging towards plausible solutions better than baselines, especially compared to the MA-WGAN that only relies on an adversarial objective. We also observe a visually more faithful orientations estimation by our method in the splenium of the corpus callosum and in the pons. In \figref{fig:gfa_comparison}, we compare the real GFA map of a test subject to the generated maps by the tested methods. We also compare the real and synthesized GFA maps to the associated HR T1w image of this said subject. It can be seen from \figref{fig:gfa_comparison} that our method generates GFA maps that are more consistent with the real HR T1w image of the subject while visually reducing boundary artifacts. Our method also seems to recover fine anatomical details that are present in the HR T1w but not in the real diffusion. Indeed, the GFA maps generated by our method exhibit sharper edges and better tissue delineation particularly at the bottom of the coronal slice where the real diffusion lacks details. We provide in \ref{appendix_visual_assessment} the color encoded FA and FA maps of five additional test subjects to visually assess the generated images by our method on different brain geometries. In \figref{fig:odf_crossing}, we compare the ODFs produced by our method and baselines in a region with crossing fibers. One can see that our method generates ODFs that smoothly transition between orientations and plausible crossing estimation.      

\subsection{Streamlines Length and Volume}
    We report in \figref{fig:results_streamline_wholebrain} the mean streamline length (in mm) and volume (in voxels) for the reconstructed wholebrain tractograms of real and generated diffusion. We further detail in \figref{fig:results_streamline_len} the mean streamline length and, in \figref{fig:results_streamline_vol}, the mean bundle volume for each segmented bundles.
    
    We can observe from the results that reconstructed whole brain tractograms are very similar in size, but that individual bundles segmented from synthesized data tend to be shorter. Indeed, looking at the mean bundle lengths and volumes reported in \tabref{tab:results_streamline_stats}, it can be seen that bundles segmented from the generated DT/ODF tractograms are respectively 14.98\% and 2.32\% shorter than real data. Nonetheless, generated ODFs tend to produce bundles that are slightly more voluminous with mean volume of $\sim$24,408 voxels compared to $\sim$24,210. While some bundles were only segmented on the real data or the generated data (IFOF vs. TPT, for example), we can observe that the bundles recovered in both cases exhibit similar statistics.
    
    \begin{table}[t!]
    \centering
    \caption{Measures on the real and generated segmented bundles from tractography.}
    \label{tab:results_streamline_stats}
    \begin{footnotesize}
    \begin{tabular}[t]{llcc}
    \toprule
    & & \textbf{Real} & \textbf{Generated}\\
    & & (mean \textpm~std) & (mean \textpm~std)\\
    \midrule
    \midrule
    \multirow{2}{*}{\textbf{DT}} & Length (mm) & 99.99 \textpm~54.37 & 85.01 \textpm~50.02\\
    & Volume (voxels) & 15963.62 \textpm~24199.55 & 14119.52 \textpm~24274.11\\
    \midrule
    \multirow{2}{*}{\textbf{ODF}} & Length (mm) & 120.94~\textpm~72.29 & 118.13~\textpm~71.57 \\
    & Volume (voxels) & 24210.36~\textpm~42082.48 & 24408.78~\textpm~44255.32\\
    \bottomrule
    \end{tabular}
    \end{footnotesize}
    \end{table}

\subsection{Streamlines Shape}
    We report in \figref{fig:results_streamline_wholebrain} the Dice, OL and OR between the wholebrain tractograms of real and generated diffusion. The same measures are then detailed in \figref{fig:results_streamline_shape} for all segmented bundles. We can see from these two figures that, despite their similarity in length and volume, the space occupied by segmented bundles from real and generated data may vary. This disparity can be observed in \figref{fig:dti_visualization} where we compare three real and generated segmented bundles. Furthermore, we observe a high inter-bundle metrics performance variability. For instance, bundles such as the Frontal Aslant Tract (FAT), Frontopontine (FPT) and Vertical Occipital Fasciculus (VOF) reach a high agreement (i.e. Dice, OL and OR close to 1) whereas the Superior Cerebellar Peduncle (SCP) and Occipitopontine Tract (OPT) are hardly matched. This high inter-bundle variability can be further appreciated in \tabref{tab:results_streamline_shape} where we note the mean Dice, OL and OR and their standard deviation for all segmented bundles.
    
    \begin{table}[t!]
    \centering
    \caption{Average metrics from the bundle shape comparison between real and synthesized data.}
    \label{tab:results_streamline_shape}
    \begin{footnotesize}
    \begin{tabular}[t]{lC{2.0cm}C{2.0cm}C{2.0cm}}
    \toprule
    & \textbf{Dice} & \textbf{OL} &  \textbf{OR}  \\
    & (mean \textpm\ std) & (mean \textpm\ std) &  (mean \textpm\ std)  \\
    \midrule\midrule
    \textbf{DT} & 0.46 \textpm\ 0.24 & 0.38 \textpm\ 0.21 & 1.36 \textpm\ 0.86 \\
    \midrule
    \textbf{ODF} & 0.54 \textpm\ 0.24 & 0.42 \textpm\ 0.19 & 1.23 \textpm\ 0.58 \\    \bottomrule
    \end{tabular}
    \end{footnotesize}
    \end{table}
    
    From \tabref{tab:results_streamline_shape}, we can also see that the segmented bundles from the generated ODF are more faithful to their real counterpart than DT with a mean Dice of 0.54 \textpm\ 0.24, a mean OL of 0.42 \textpm\ 0.19 and a mean OR of 1.23 \textpm\ 0.58.

\section{Discussion}\label{sec:discussion}

    In this work, we propose a novel Riemannian deep learning architecture for the synthesis of 3D manifold-valued data and have tested its performance on two tasks: 1) the generation of diffusion tensors (DTs) and 2) the generation of diffusion orientation diffusion functions (ODFs). Specifically, we have explored the feasibility of generating high-resolution DT and ODF from high-resolution structural T1w images and unpaired LR diffusion. We show in \tabref{tab:non_manifold_count} that a standard model relying on Euclidean operations fails to capture the geometry of the diffusion data manifold which leads to the estimation of physically incorrect diffusion. To alleviate this issue, we have built a framework on top of recent advances in manifold-valued data processing and Riemannian geometry \citep{Arsigny2006, Cheng2009,Huang2019} to ensure the validity of the generated diffusion. We have evaluated the generated volumes properties using mean squared errors of FA/GFA maps, geodesic distances and cosine similarities between real and predicted principal fiber orientation. To further evaluate the integrity of the synthesized diffusion in a typical diffusion application, we have performed tractography and assessed the lengths, volumes and shapes of resulting tractrograms.  
    
    %%\subsection{Image Prior Regularization}
    %%To stabilize the training of our model and to ensure the convergence towards solution with expected properties, we introduced paired samples from a minimal number of training subjects. We observed a constant increase of performance in all metrics when sampling paired patches from up to 50 subjects as reported in \tabref{tab:paired_subjects_results} followed by a small reduction when using 75 subjects. One could have expected a monotonic increase of performance inline with the number of subjects but here, for computational limitations, we also limited the number of sampled patches to 50\,000 in all scenarios. Consequently, the sampling sparsity per subject increases with the enlargement of the paired subjects pool. We observed the best performance for all models when using paired patches from 50 different subjects which correspond to roughly 1000 patches per subject. The optimal sampling of large image datasets is still an open question and is out of the scope of this present work. However, it seems that a good balance between subject diversity and per-subject sampling density is beneficial.
    
    \subsection{Diffusion Synthesis Performance}
        The generation of DT/ODF solely relying on a T1w image is an ill-posed problem for which a single T1w intensity can correspond to several fiber arrangements. However, by providing the contextual information required for the network to localize the structural input, we observe that strong fiber patterns can successfully be recovered by our method and baselines. Hence, we believe that, to a certain extent, the high-level geometry of the brain globally impacts its underlying fibers organization. This seems to be particularly true in regions of higher anisotropy where fiber tracts are strongly organized. As a result, we observed a better estimation of the principal fibers orientation in regions with higher FA/GFA and a generally poor estimation of the principal orientations in regions with high inter-subject variance such as the ventricles.
        
        Using HR structural images to drive the synthesize of diffusion helps recovering fine anatomical details and sharp edges better than interpolation based methods. By leveraging the detailed information contained in HR structural images, our network is not limited by the coarseness of the low-resolution input diffusion signal such as in interpolation.
        
        This transfer of information from HR to LR images is enforced by our cycle-consistency loss that preserves a high structural coherence between the HR structural inputs and the generated diffusion. In addition, our adversarial and prior loss help recovering plausible fiber patterns better than our baselines by leveraging unpaired examples of real diffusion. Furthermore, the manifold-awareness yields on-par or better metrics when compared with equivalent Euclidean architectures. The manifold-awareness, however, comes with the benefit of ensuring the mathematical properties of the synthesized diffusion regardless of the amount of network training. 

\subsection{Tractography Performance}

    \subsubsection{DT vs. ODF}
        We observe from \tabref{tab:results_streamline_stats} that bundles segmented from ODF tractography tend to be longer and more voluminous which is an expected behavior. While a thorough comparison between DT and ODF tractography is out-of-scope for this work (c.f. \citep{Farquharson2013, Thomas2014, Jeurissen2019}), we can nevertheless appreciate that the synthesized DT and ODF behave in a manner similar to their real counterparts in the context of tractography.
        
        Moreover, we observed that the bundles segmented from ODF tractography have a higher mean Dice, higher overlap and lower overreach than bundles segmented from DT tractography. Since the same algorithm was used to perform tractography on both datasets, the main difference is that DT tractography makes use of a single direction while ODF tractography may use multiple local maxima to propagate streamlines. As such, the lower discrepancies in ODF bundle shapes can be explained by the multiple directions used in tractography compensating for local errors. At the opposite, DT tractography is known to be sensitive to local estimation error \citep{Huang2004}. This sensitivity, which often lead to the early termination or to the switch to a wrong adjacent tract of the tracking algorithm \citep{Jeurissen2019}, can greatly affect the final tractograms shape.  
    
    \subsubsection{Bundle shape analysis}
        Since DT/ODF generation studies are still few, it is hard to provide a definitive conclusion on the quality of the streamlines generated on synthesized data. However, we observe from \figref{fig:results_streamline_wholebrain} that whole brain tractograms have a similar streamline length, occupy the same number of voxels and have the same shape. Segmented bundles, if extracted from both real and generated data, also tend to exhibit the same average length and volume.
        
        While the reported bundle shape metrics might seem to indicate a poor correspondence between bundles generated from real and synthesized data, it should be noted that bundle segmentation is a highly variable operation. For example, \citet{Rheault2020}, which analyzed the reproducibility of the segmentation of a single bundle between human experts and non-experts, reports a significant difference between the volume of segmented bundles between experts and non-experts as well as median Dice scores around 0.77 for intra-rater reproducibility, 0.65 for inter-rater reproducibility and 0.8 for reproducibility with a gold standard. In a similar manner, \citet{Schilling2020} analyzed the variability in the segmentation of 14 bundles between 42 groups using both manual and automatic segmentation. While few actual metrics are reported, figures indicate a generally low Dice score, as well as a high variability in bundle volume and streamline length for inter-protocol responsibility. Analysis for specific pathways report Dice scores between 0.4 and 0.6 for inter-protocol and inter-subject reproducibility and Dice scores between 0.6 and 0.8 for intra-protocol reproducibility. As such, we can theorize that the reported Dice scores in the present work could be impacted by the inherent variability in the segmentation process.
        
\section{Conclusion}\label{sec:conclusion}
    This work presents a novel Riemannian network architecture for the cycle-consistent synthesis of diffusion tensors and diffusion ODF in high-resolution structural T1w space. The results have demonstrated that our Riemannian architecture can synthesize valid diffusion images with a 5\% improvement in principal fibers orientation and a 23\% improvement in FA MSE with respect to our baselines. The better performance of our approach over compared methods shows the benefit of using both paired and unpaired samples in a single objective. Furthermore, as opposed to standard Euclidean deep learning models, which generate an average of 3,844 invalid tensors and 1,560 invalid ODFs per volume, our method guarantees the mathematical coherence of the synthesized diffusion schemes, free of invalid tensors or ODFs. 
    
    Moreover, we have evaluated qualitatively our generated diffusion volumes by comparing their tractograms with their real counterparts. It was observed that our generated T1w-driven diffusion shares similarities with the real diffusion in terms of streamline length, volume and fiber bundles shape. We have also shown the ability of our network to transfer fine anatomical details from the high-resolution T1w images to diffusion images. This transfer of information allows the generation of images with sharper edges and a higher level of details that could not be achieved with image interpolation.   
    
    Our results suggest that the high-level geometry of the brain, encoded in structural T1w images, can be used to predict its global fiber bundles organization. Leveraging this principle, our method could enable the fast synthesis of DT and ODF in situations where the acquisition of diffusion imaging is not available. More generally, it offers the basis of a framework targeting any real-to-manifold-valued image translation tasks. For instance, our method could be used for missing modalities synthesis and datasets completion, manifold-valued image inpainting or manifold-valued population based statistics that rely on non-Euclidean metrics.

\section*{Acknowledgments}
This work was supported financially by the Canada Research Chair on Shape Analysis in Medical Imaging, the Research Council of Canada (NSERC), the Fonds de Recherche du Québec (FQRNT), the Réseau de Bio-Imagerie du Québec (RBIQ), and ETS Montreal.

%%Harvard
\bibliographystyle{model2-names.bst}\biboptions{authoryear}
\bibliography{sample}

\appendix

\section{Backpropagation for Diffusion Tensors Learning} \label{appendix_backprop}
The $\logId(\,\cdot\,)$ and $\expId(\,\cdot\,)$ maps ensure that our generator network synthesizes valid diffusion tensors. Both maps involves the EIG operator on $3\times3$ symmetric matrices whose gradients must be defined to train the network with the standard backpropagation algorithm. Employing the matrix generalization of backpropagation \citep{Ionescu, Huang2016}, we define the partial derivatives of the objective function at the $k^{th}$ layer with respect to a generated diffusion tensor $\matr{M}$ as follows:
\begin{equation}\label{eq:derivative_spectral_layer}
\footnotesize
\forall \matr{M}_{k-1} \in \matr{X_{k-1}}, \dLdM = 2\UU \bigg(\tr{\KK} \circ \sym{\bigg(\tr{\UU}\dLdU\bigg)}\bigg)\tr{\UU}\bigg) + \UU \diag{\bigg( \dLdE\bigg)}\tr{\UU},
\end{equation}
\noindent with
\begin{equation}\label{eq:eigen_decomposition}
   \matr{M}_{k-1} = \UU \Eig \tr{\UU},
\end{equation}
\noindent and
\begin{equation}\label{eq:matrix_k}
    \KK(i,j) = \left\{\begin{matrix}
\frac{1}{\sigma_{i} - \sigma_{j}} & i\neq j \\ 
0 & i = j,
\end{matrix}\right.
\end{equation}
\noindent where $\sym{\matr{A}}= \frac{1}{2}(\matr{A} + \tr{\matr{A}})$ and $\diag{\matr{A}}$ is $\matr{A}$ with nonzero elements only in its diagonal. Furthermore, $\matr{X_{k-1}} \in \real^{B \times 9 \times D \times H \times W}$ is a batch of 3D DT patches with size $D \times H \times W$ and $\matr{K}$ is built upon the eigenvalues $\sigma$ in $\matr{\Sigma}$ of $\matr{M}$.
\bigbreak
The variations $\dLdU$ and $\dLdE$ in \equaref{eq:derivative_spectral_layer} are defined for every tensor $\matr{M}$ for both maps $\logId(\cdot)$ and $\expId(\cdot)$ respectively as:
\begin{equation}\label{eq:derivative_U_log}
\dLdU = 2\sym{\bigg(\dLdMk\bigg)}\UU \log(\Eig),
\end{equation}
\begin{equation}\label{eq:derivative_E_log}
\dLdE = \Eig^{-1}\tr{\UU}\sym{\bigg(\dLdMk\bigg)}\UU,
\end{equation}
\noindent and
\begin{equation}\label{eq:derivative_U_exp}
\dLdU = 2\sym{\bigg(\dLdMk\bigg)}\UU \exp(\Eig),
\end{equation}
\begin{equation}\label{eq:derivative_E_exp}
\dLdE = \exp(\Eig)\tr{\UU}\sym{\bigg(\dLdMk\bigg)}\UU.
\end{equation}

With the help of \equaref{eq:derivative_spectral_layer} and the successive application of the $\expId(\cdot)$ and $\logId(\cdot)$ maps, we make sure that the generated tensors are and stay SPD throughout the training. An open-source implementation of these differentiable functions is available on our Github \footnote{https://github.com/banctilrobitaille/torch-vectorized}\footnote{https://torch-vectorized.readthedocs.io/en/latest/}.

\onecolumn

\section{Tractography Assessment}\label{appendix_tractography_assessment}

\begin{figure}[hbt!]
    \centering
    \includegraphics[width=0.97\linewidth]{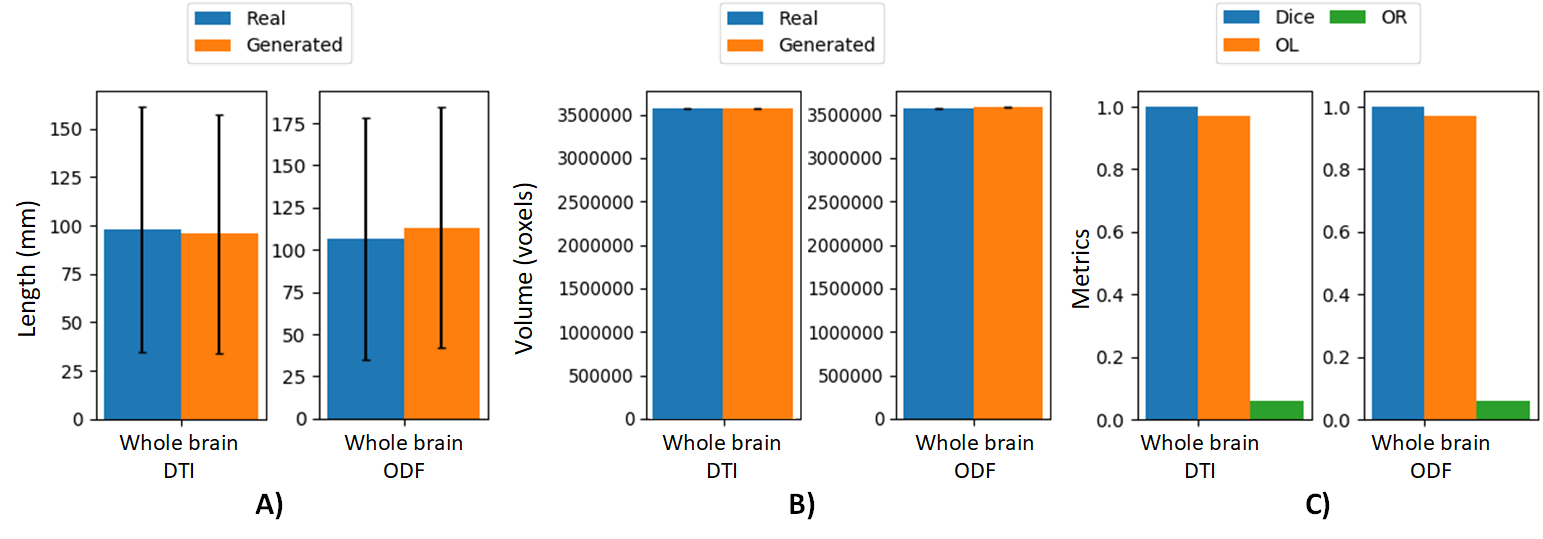}
    \caption{In \textbf{A)}, the mean streamline length (in mm). In \textbf{B)}, the mean volume (in voxels) and in \textbf{C)}, the mean Dice, OL and OR.} \label{fig:results_streamline_wholebrain}
    \end{figure}

    This appendix presents additional figures, referenced in \secref{sec:experiments} and in \secref{sec:discussion}, that compare the generated and the expected real tractograms. First, we compare in \figref{fig:results_streamline_wholebrain} the mean volume, streamline length, Dice, OL and OR when considering the whole brain tractograms.   

    We then compare the mean volume of each recovered bundle in \figref{fig:results_streamline_vol} below: 
    
    \begin{figure}[hbt!]
    \centering
    \includegraphics[height=0.43\textheight]{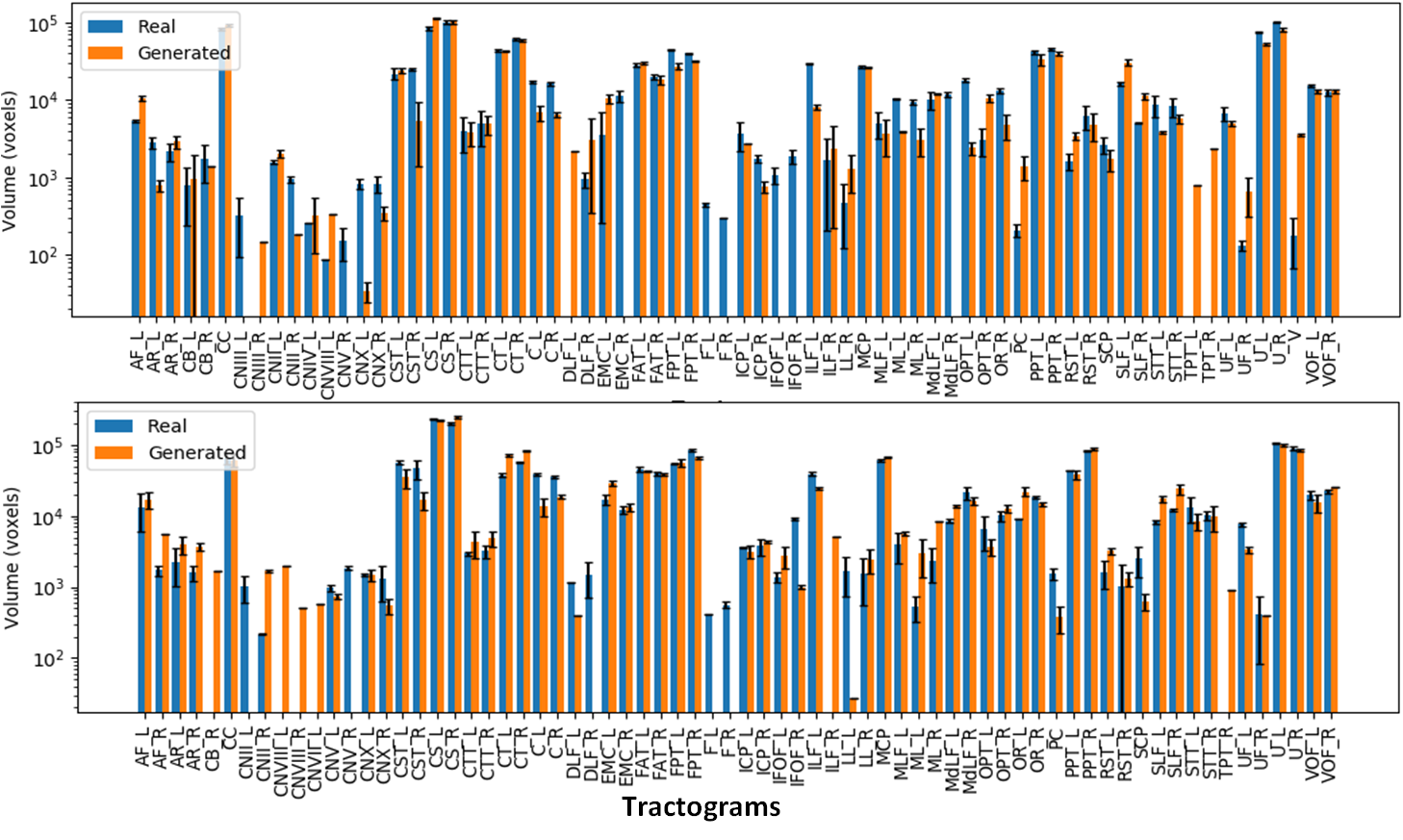}
    \caption{\textbf{Top row}: Mean bundle volumes (in voxels) from real and generated DT. \textbf{Bottom row}: Mean bundle volumes (in voxels) from real and generated ODF.} \label{fig:results_streamline_vol}
    \end{figure}
    
\begin{figure}[hbt!]
    \centering
    \includegraphics[height=0.43\textheight]{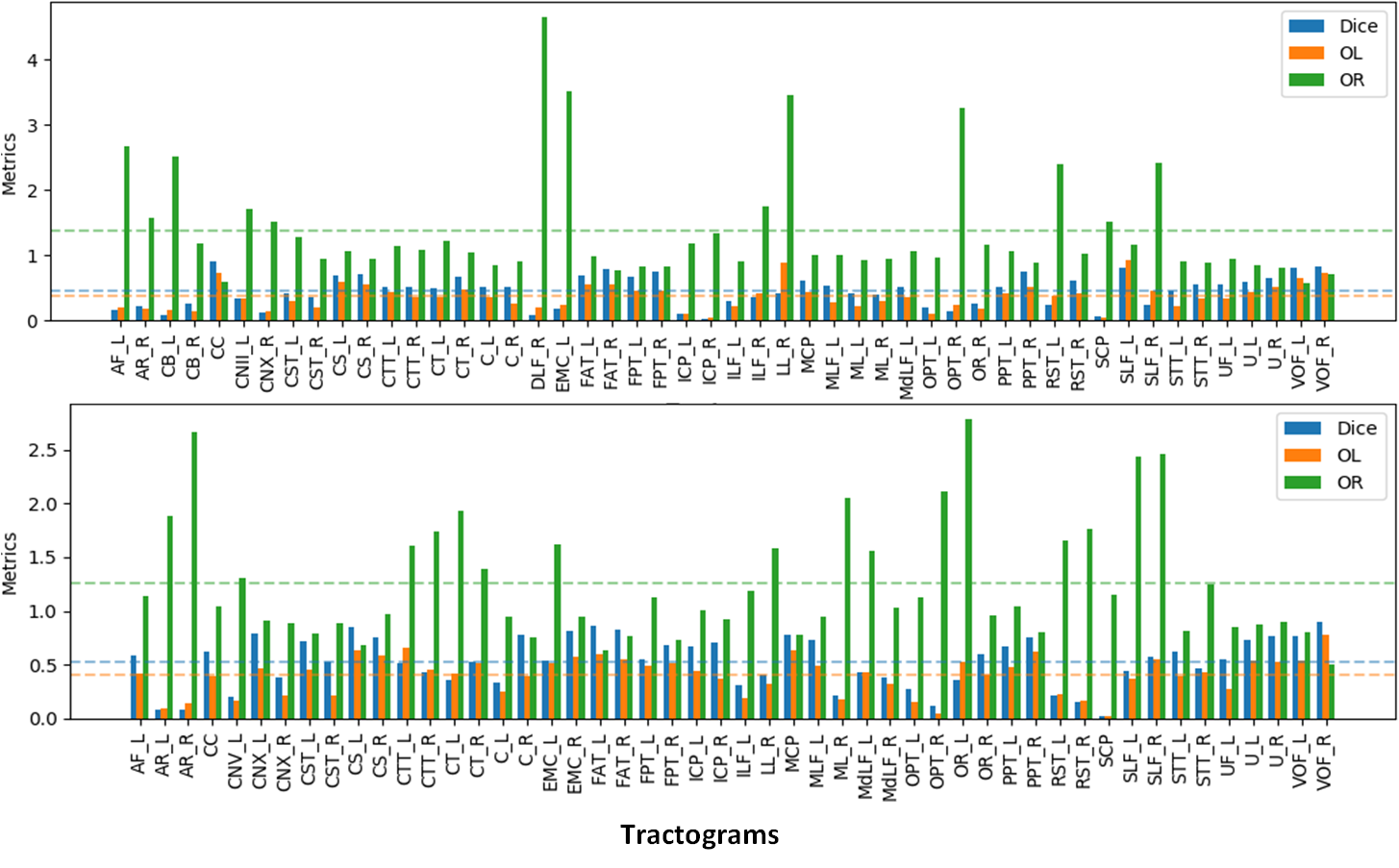}
    \caption{\textbf{Top row}: Dice, OL and OR between bundles and tractograms from real and generated DT. \textbf{Bottom row}: Dice, OL and OR between bundles and tractograms from real and generated ODF. The lines indicate the mean metric values for all bundles.} \label{fig:results_streamline_shape}
    \end{figure}

    Finally, \figref{fig:results_streamline_shape} presents the Dice, OL and OR yielded by each recovered bundles.
    
\clearpage

\section{Visual Assessment of Synthesized Volumes}\label{appendix_visual_assessment}
    
\begin{figure*}[hbt!]
    \centering
    \includegraphics[height=0.37\textheight]{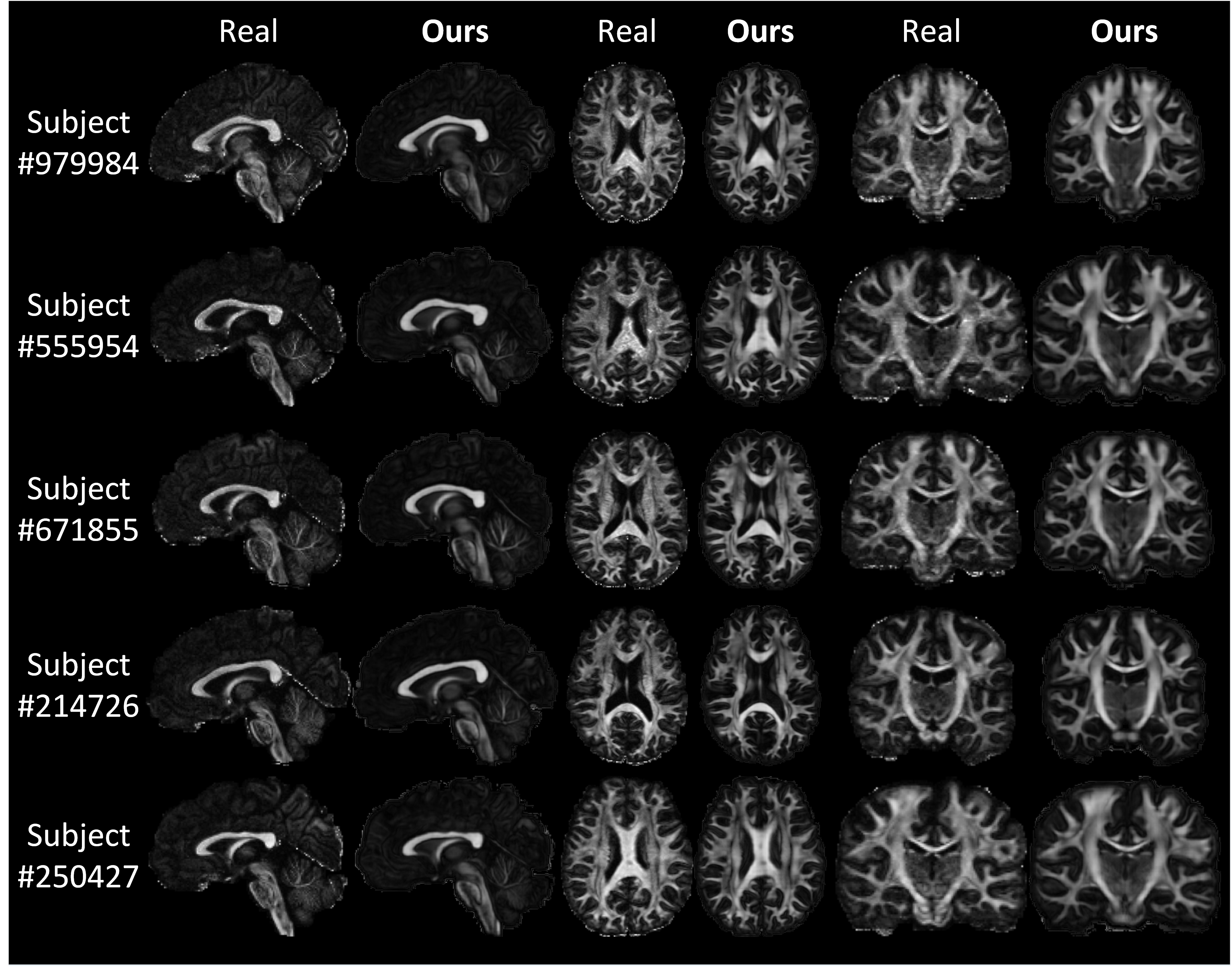}
    \caption{The real and synthesized FA maps of five test subjects in the sagittal, axial and coronal planes.}\label{fig:appendix_multi_subjects_FA}
    \end{figure*}

\begin{figure*}[hbt!]
    \centering
    \includegraphics[height=0.37\textheight]{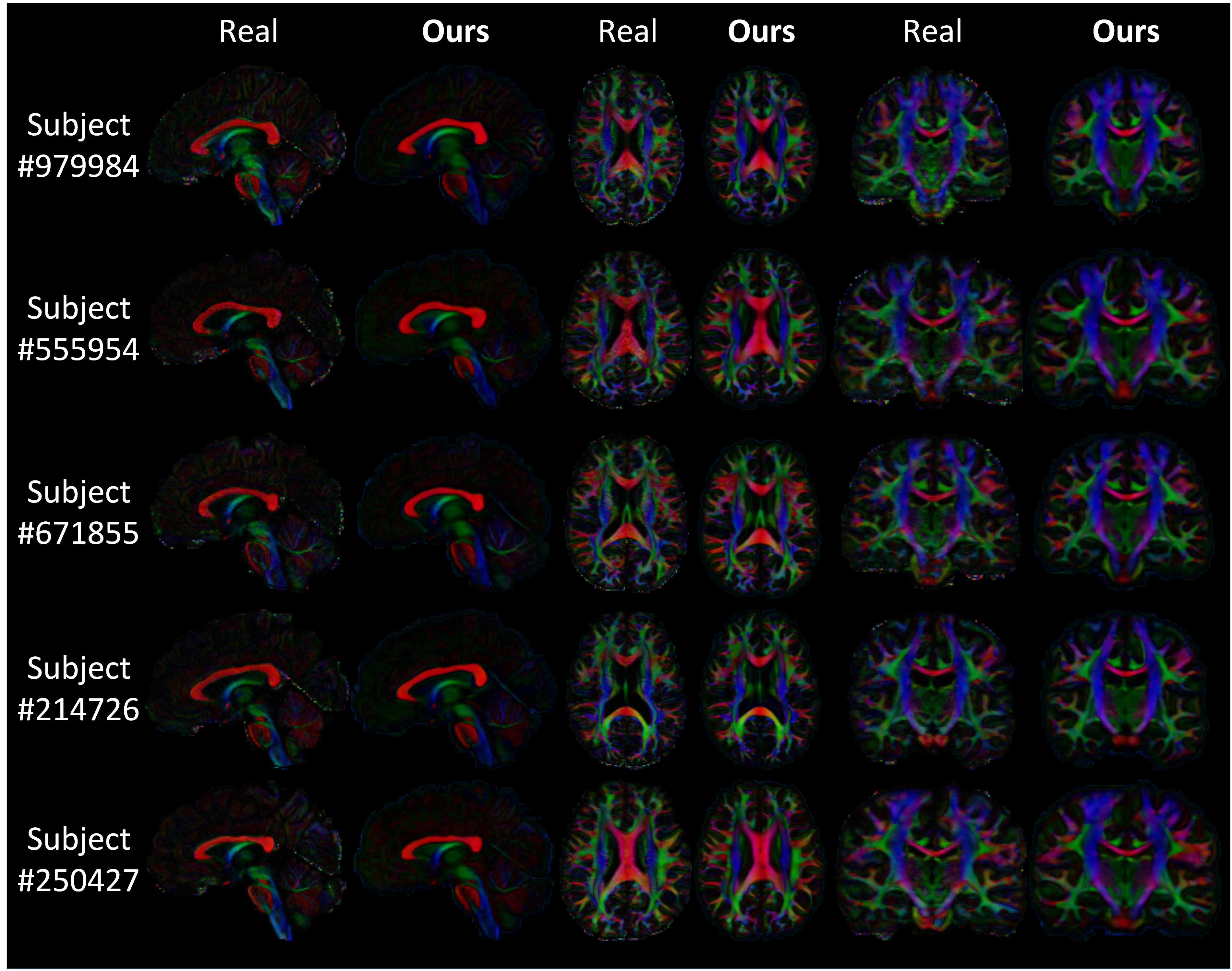}
    \caption{The real and synthesized color FA maps of five test subjects in the sagittal, axial and coronal planes.} \label{fig:appendix_multi_subjects_CFA}
    \end{figure*}

    We provide, in the following appendix, the FA and color FA maps of five additional test subjects. We first compare in \figref{fig:appendix_multi_subjects_FA} the generated and real FA of a sagittal, axial and coronal slice. We then compare in \figref{fig:appendix_multi_subjects_CFA} the generated and real color FA of the same five subjects. One can appreciate how our method adapts to different brain geometries and generates plausible FA and principal orientations despite the differences in the anatomy of the subjects. 
    
\end{document}